\pdfoutput=1

\documentclass[11pt]{article}

\usepackage{acl}

\usepackage{times}
\usepackage{latexsym}

\usepackage[T1]{fontenc}

\usepackage[utf8]{inputenc}

\usepackage{microtype}

\usepackage{inconsolata}
\usepackage{rotating}
\usepackage{enumitem}
\usepackage{amsmath}
\usepackage{amsthm}
\usepackage{tikz}
\usepackage{pgfplots}
\usepackage{pgfplotstable}
\usepackage{booktabs}
\usepackage{amssymb}
\usepackage{adjustbox}
\usepackage{makecell}
\usepackage{scalerel}
\usepackage{comment}
\usepackage{svg}
\usepackage{xspace}
\usepackage{subcaption}
\usepackage[linesnumbered,ruled,vlined]{algorithm2e}
\usepackage{float} 
\usepackage{algorithmic}

\usepackage{cleveref}
\crefname{section}{\S}{\S\S}
\Crefname{section}{\S}{\S\S}
\crefformat{section}{\S#2#1#3}
\crefname{appendix}{App.}{}
\crefname{figure}{Fig.}{Fig.}
\crefname{table}{Table}{Tables}
\crefname{equation}{eq.}{eqs.}
\crefname{algorithm}{Alg.}{Algs.}
\creflabelformat{equation}{#2\textup{#1}#3}

\usepackage{tabularray}
\usepackage{cellspace}
\usepackage{setspace}
\makeatletter
\DeclareRobustCommand*{\escapeus}[1]{%
    \begingroup\@activeus\scantokens{#1\endinput}\endgroup}
\begingroup\lccode`\~=`\_\relax
\lowercase{\endgroup\def\@activeus{\catcode`\_=\active \let~\_}}
\makeatother
\usepackage[scaled=0.85]{helvet}

\theoremstyle{definition}
\newtheorem{example}{Example}[section]

\allowdisplaybreaks[4]
\newcommand{\ie}{\textit{i.e.}}
\newcommand{\eg}{\textit{e.g.}}

\newcommand{\finewebtwo}{FineWeb2}

\newcommand{\thirtylangs}{\textit{30-lang}}
\newcommand{\sixtylangs}{\textit{60-lang}}
\newcommand{\pbpe}{Parity-aware BPE}
\newcommand{\unbalanced}{\textcolor{BrickRed}{\textit{unbalanced}}\xspace}
\newcommand{\balanced}{\textcolor{OliveGreen}{\textit{balanced}}\xspace}

\newcommand{\defn}[1]{\textbf{#1}}
\newcommand{\token}{v}
\newcommand{\tokensequence}{\mathbf{\token}}
\newcommand{\vocab}{\mathcal{V}}
\newcommand{\byte}{b}
\newcommand{\bytesequence}{\mathbf{\byte}}
\newcommand{\bytevocab}{\mathcal{B}}
\newcommand{\merge}{m}%
\newcommand{\merges}{\mathbf{\merge}}
\newcommand{\corpus}{\mathcal{D}}
\newcommand{\tokenizer}{T}

\newcommand{\tok}{\tau}
\newcommand{\detok}{\ensuremath{\rotatebox[origin=c]{180}{$\tok$}}}
\newcommand{\comprate}{\mathrm{CR}}
\newcommand{\defeq}{\mathrel{\stackrel{\textnormal{\tiny def}}{=}}}
\newcommand{\languages}{\mathcal{L}}
\newcommand{\lang}{\ell}
\newcommand{\multilingcorpus}{\mathcal{M}}
\newcommand{\devcorpus}{\corpus^{\scaleto{\mathrm{dev}}{4pt}}}

\definecolor{darkgreen2}{HTML}{009B55}
\definecolor{myblue2}{HTML}{29abe2}
\definecolor{myorange2}{HTML}{f7931e}

\definecolor{mypurple2}{HTML}{9823FF}

\title{Parity-Aware Byte-Pair Encoding:\\ Improving Cross-lingual Fairness in Tokenization}

\author{
    Negar Foroutan\textsuperscript{1}\thanks{Equal contribution, \textdagger Equal supervision.} \qquad
    Clara Meister\textsuperscript{1}\footnotemark[1] \qquad
    Debjit Paul\textsuperscript{1} \\
    \textbf{Joel Niklaus\textsuperscript{2}} \qquad
    \textbf{Sina Ahmadi\textsuperscript{3}} \qquad
    \textbf{Antoine Bosselut\textsuperscript{1}\textsuperscript{\textdagger} }  \qquad
    \textbf{Rico Sennrich\textsuperscript{3}\textsuperscript{\textdagger}}\\
\\
\textsuperscript{1}EPFL \quad
\textsuperscript{2}Niklaus.ai \quad
\textsuperscript{3}University of Zurich \quad \\ 
}

\begin{document}
\maketitle

\begin{abstract}
Tokenization is the first---and often least scrutinized---step of most NLP pipelines. 
Standard algorithms for learning tokenizers rely on frequency-based objectives, which favor languages dominant in the training data and consequently leave lower-resource languages with tokenizations that are disproportionately longer, morphologically implausible, or even riddled with \texttt{<UNK>} placeholders.
This phenomenon ultimately amplifies computational and financial inequalities between users from different language backgrounds. 
To remedy this, we introduce Parity‑aware Byte Pair Encoding (BPE), a variant of the widely-used BPE algorithm. At every merge step, Parity‑aware BPE applies a fair-max rule that maximizes the compression gain of the currently worst‑compressed language, trading a small amount of global compression for cross‑lingual parity. 
We find empirically that Parity-aware BPE reduces tokenization inequality---operationalized by the Gini coefficient of per-language token costs---by up to 89\% relative to Classical BPE. This comes with negligible impact on global compression rate and no evidence of systematic degradation in downstream LM performance.\footnote{Code for reproducing experiments is available \href{https://github.com/swiss-ai/parity-aware-bpe.git}{here}; Parity-aware BPE is also available in HuggingFace.}

\end{abstract}

\section{Introduction}

At a time of rapid innovation and constant change in natural language processing (NLP), tokenization remains a foundational and comparatively stable component of NLP pipelines. 
Tokenization is the process of transforming raw sequences of bytes\footnote{Early work considered characters the ``base unit'' of strings, but raw bytes have become popular
because their fixed 256‑symbol vocabulary can encode any character from any encoding, eliminating out‑of‑vocabulary issues.
} into sequences of byte-spans, \ie, subwords; it enables computational efficiency and provides essential inductive biases by defining meaningful textual units. 
This design choice can have a major impact on various aspects of model performance \citep{bostrom-durrett-2020-byte,ali-etal-2024-tokenizer,goldman-etal-2024-unpacking}.\looseness=-1

The predominant tokenization algorithms---such as Byte Pair Encoding \citep[BPE;][]{sennrich-etal-2016-neural} and UnigramLM \citep{DBLP:conf/acl/Kudo18}---construct the vocabulary by maximizing frequency-based objectives computed over an entire training corpus. 
In multilingual corpora, this global criterion inevitably favors the languages with the greatest representation.
Under vocabulary size constraints, subwords that primarily benefit high-resource languages are preferentially included, often at the expense of those needed for lower-resource languages.
This bias has both qualitative and economic consequences. 
Models trained on fragmented or semantically incoherent tokenizations lose valuable inductive biases and tend to perform worse. Simultaneously, texts in lower-resource
languages---often tokenized into more tokens---incur higher computational costs from LLM-based services charging based on token count.
This ``tokenization tax'' disproportionately burdens users of underrepresented languages and exacerbates existing inequalities.

In an effort to mitigate these inequities, we introduce \textbf{Parity-aware BPE}. 
The classic BPE algorithm constructs its vocabulary by iteratively selecting the subword pair with the highest corpus-level frequency; it adds the concatenated pair to the vocabulary and replaces all instances of the pair with the new symbol.\footnote{This process acts as a form of data compression, replacing frequent subword sequences with shorter representations.} 
Parity-aware BPE is a simple variant of this algorithm, retaining the iterative framework but redefining the merge selection rule: at each step, it computes co-occurrence statistics separately for each language and then uses statistics from the language with the current worst compression rate for selecting the next merge. 
In other words, instead of greedily maximizing a global objective, Parity-aware BPE performs a ``fair-max'' update that progressively equalizes string compression rates across languages. Notably, this modification affects only the vocabulary learning phase; the inference procedure remains the same as in Classical BPE.\looseness=-1

Empirically, Parity-aware BPE substantially improves cross-lingual tokenization fairness. In our setup that mimics standard LM training settings, we observe an 89\% reduction in the Gini tokenizer inequality coefficient compared to Classical BPE. 
Across 12 multilingual benchmarks, models trained with a Parity-aware BPE tokenizer match downstream performance compared to those trained with a Classical BPE tokenizer. 
In summary, Parity‑aware BPE reduces tokenizer disparities between languages while showing no systematic degradation in downstream performance relative to Classical BPE.

\section{Text Tokenization}\label{sec:tokenization_background}

Text can be decomposed at many granularities: graphemes, Unicode code points, or multi-character tokens. At the lowest level of abstraction, however, the digital representation of a string is simply a sequence of \defn{bytes}---the foundation on which all other units are constructed. 
Let $\byte \in \bytevocab = \{0,\dots,255\}$ denote an individual byte.  
A finite \defn{byte-string} is defined as $\bytesequence \in \bytevocab^*$, where $\bytesequence = b_{1} b_{2} \cdots b_{|\bytesequence|}$.  
While we treat bytes as the base unit throughout this paper, the following definitions remain compatible with alternative base units, such as characters or graphemes.

\subsection{Byte-level Tokenizers}
In plain terms, \defn{tokenization} is the process of mapping raw byte-strings to sequences of subwords. 
A \defn{tokenizer} specifies the rules that perform this mapping.
Tokenizers consist of two components: a \defn{vocabulary}  $\vocab \subset \bytevocab^{+}$---a finite set of non-empty byte-spans, often called \textit{subwords}\footnote{To guarantee representability of any byte-string, we assume $\vocab$ includes all singleton bytes: $\bytevocab \subseteq \vocab$.}---and a  \defn{tokenization function} $\tok: \bytevocab^* \!\to\! \vocab^*$---a mapping from byte-strings
  $\bytesequence$ to sequences of tokens
  $\tokensequence = \token_{1},\token_{2}, \ldots$.

\paragraph{Pre-tokenization and Normalization.}

Many tokenization algorithms include a \emph{pre-tokenization} (and often a normalization) step that segments or rewrites raw byte strings according to deterministic criteria. 
Pre-tokenization can encompass several operations, including Unicode normalization or whitespace splitting. 
Notably, pre-tokenization determines subword boundaries, and thus also determines the set of possible candidates for the vocabulary as well as the attainable compression rate. For example, if whitespace is used as a subword boundary, languages without explicit whitespace (\eg, Chinese, Japanese) or with rich morphology may have the potential for higher compression. 
While often overlooked, pre-tokenization choices have a large impact on a tokenizer's learned vocabulary; we point the interested reader to \citep{arnett2025crosslingual}. 
For simplicity, we assume the pre-tokenization step is baked into $\tok$. 

\subsection{Text Compression}

\textit{Why does mapping a raw byte‑string to a sequence of larger subword tokens help an LM?} 
While byte- or character-level sequences can be provided directly to an LM, subword tokenization remains the dominant approach for turning text into model inputs. 
The precise inductive biases this process imbues remain an open research question \citep{zouhar-etal-2023-tokenization,schmidt-etal-2024-tokenization},
but one plausible explanation is the \emph{compression} it provides: a good tokenizer tends
to map each input $\bytesequence$ to a shorter sequence of tokens, reducing the length of the model's effective input and, potentially making learning easier. At the very least, it can significantly reduce computational overhead.

For a fixed tokenizer $(\vocab,\tok)$ we define the
\defn{compression rate} of a byte‑string $\bytesequence$ as
\vspace{-2mm}
\begin{equation}\label{eq:compression}
     \comprate(\bytesequence;\tok)
    \;\defeq\;
    \frac{\lvert\bytesequence\rvert_u}{\lvert\tok(\bytesequence)\rvert}
\end{equation} 

\noindent where $|\bytesequence|_u$ denotes the length of $\bytesequence$ in terms of a given \defn{normalization unit} $u$ (\eg, characters, words, lines, or simply bytes). 
In words, $\comprate(\bytesequence;\tok)$ measures the factor by which our original sequence length is reduced after tokenization.
A higher $\comprate$ indicates stronger compression. 

We are generally interested in a tokenizer's average compression, which can be estimated from a corpus $\corpus$:

\begin{equation}\label{eq:comp_rate}
     \comprate(\corpus;\tok)
    \;\defeq\;
\frac{\sum_{\bytesequence\in\corpus}\lvert\bytesequence\rvert_u}{\sum_{\bytesequence\in\corpus}\lvert\tok(\bytesequence)\rvert}
\end{equation}

This quantity provides an estimate of the average number raw text units $u$ that are packed into each token that an autoregressive LM processes. 
Tokenizers differ in how small they can make $\mathrm{CR}(\corpus;\tok)$ while remaining lossless. 
Further, this rate can vary across strings from different languages, which motivates our last definition: language-specific compression rate.\looseness=-1 

Let $\languages=\{\lang^{(r)}\}_{r=1}^{R}$ be our set of languages and $\multilingcorpus=\{(\bytesequence^{(s)},\lang^{(s)})\}_{s=1}^{S}$ be a labeled multilingual corpus, \ie, a corpus where each byte‑string $\bytesequence^{(s)}$ is labeled with its language $\lang^{(s)}$. 
For a fixed tokenizer, we define a language's compression rate as 
\begin{equation}
    \comprate(\lang; \tok)\defeq \comprate(\corpus_\lang;\tok)
\end{equation}
where $\corpus_\lang = \{\bytesequence^{(s)}: (\bytesequence^{(s)},\lang^{(s)})\in\multilingcorpus, \,\lang^{(s)}=\lang\}$.

\subsection{Tokenization Fairness} Many deployed systems bill users per token; additionally, inference latency typically scales with token count. We therefore define tokenizer cost for language $\lang$ as the expected number of tokens required to encode a piece of content in $\lang$. To make ``content'' comparable across languages, we measure cost using a parallel corpus and measure tokens per aligned segment (line/sentence/document). Inequality in these per-language costs corresponds to unequal user cost/latency purely due to language choice.  
We adopt the Gini coefficient across languages (defined in \cref{sec:intrinsic}) as our primary fairness metric and success criterion, as it captures exactly this inequality. Intuitively, this metric can be thought of as a measure of how disparate $\comprate(\lang; \tok)$ is across $\ell\in\languages$. 
More details are given in \cref{sec:intrinsic}.

\section{Byte Pair Encoding}
\textit{Here we provide a detailed description of the classic version of BPE; readers already familiar with the algorithm can skip this subsection.} 

Byte Pair Encoding \citep[BPE;][]{sennrich-etal-2016-neural} is one popular algorithm for creating a tokenizer adapted from the
byte-pair compression scheme of \citet{Gage1994}. 
In short, BPE tokenizes text by iteratively \emph{merging} adjacent tokens whose token-types (\ie, subwords) were observed to co-occur frequently in the training data. 

The notion of a \defn{merge} lets us formalize this procedure.  
A merge is defined as an ordered pair
$\merge=(v,v')$ with $v,v'\in\vocab$. 
The application of a merge to a token sequence replaces every bigram token
$v,v'$ by a single token $v\circ v'$. 
Each replacement shortens the token sequence by exactly one token, thereby \emph{compressing} the sequence. 
To tokenize a piece of text with a BPE tokenizer, we start from its representation as a byte-string, \ie, a sequence of base bytes, all of which necessarily appear in our tokenizer's vocabulary. We then iteratively apply a given list of merges $\merges$ to that sequence, in the order in which they appear in $\merges$. We denote this process with the function $\tok_{\merges}$, \ie, the tokenization function of a BPE tokenizer parameterized by $\merges$. The vocabulary of this tokenizer is then $\vocab = \bytevocab \cup \{v\circ v':(v,v') \in \merges\}$.  
Note that because the merge list is fixed in advance, the encoding is deterministic.  
Intuition for the merge procedure is perhaps best acquired by a small example: 

\begin{example}[Example of the iterative application of merge sequence $\merges$ to byte sequence $\bytesequence$ ]
\label{ex:bpe}
\begingroup\small
\[
\begin{aligned}
\merges = &[(\texttt{b},\texttt{a}),(\texttt{ba},\texttt{b})];\quad \bytesequence = \texttt{babab}\phantom{a}  \\[2pt]
&\quad\tokensequence_0 = \texttt{b},\texttt{a}, \texttt{b}, \texttt{a}, \texttt{b}\\
\text{Step 1:}\, \,& \texttt{b,a}\!\to\!\texttt{ba} 
  \,\,\Longrightarrow\,\, \tokensequence_1 =\texttt{ba}, \texttt{ba}, \texttt{b} \\ 
\text{Step 2:}\,\, & \texttt{ba,b}\!\to\!\texttt{bab} 
  \,\,\Longrightarrow\,\, \tokensequence_2 =\texttt{ba}, \texttt{bab} 
\end{aligned}
\]
\endgroup

\end{example}

\paragraph{Learning $\merges$.}
The BPE algorithm seeks the merge list $\merges^*$ (subject to a size constraint $K$) that maximizes the compression rate of the corpus $\corpus$:\looseness=-1
\vspace{-3mm}
\begin{equation}\label{eq:bpe_objective}
    \merges^*\;=\; \max_{\merges: |\merges| = K}\,\comprate(\corpus; \tok_\merges)
\end{equation}

BPE takes a greedy approach to choosing $\merges$, finding an approximate solution to \Cref{eq:bpe_objective} \citep{zouhar-etal-2023-formal}.  
It starts with the singleton‑byte vocabulary $\vocab_{0}=\bytevocab$ and repeatedly greedily enlarges the vocabulary. 
At each of $K$ steps, the current tokenizer $\tok_{\merges_{<k}}$ is applied to the entire training corpus, and the algorithm counts how often every adjacent pair of tokens occurs. 
The subword-type pair with the highest count, which we denote as $(v^{\star},v'^{\star})$, is deemed the most ``compressive.''
Its concatenation $v^{\star}\circ v'^{\star}$ is added to the vocabulary, the merge $\merge_{k}=(v^{\star},v'^{\star})$ is recorded, and every occurrence of the bigram $(v^{\star},v'^{\star})$ in the corpus is replaced by the new token so the next iteration works with updated token sequences. 
Repeating this process $K$ times yields the ordered list $\merges=[\merge_{1},\ldots,\merge_{K}]$ and the final vocabulary $\vocab_{K}$. When encoding a new text, $\tok_{\merges}$ simply applies these merges in the same order. The algorithm pseudocode is provided in \cref{alg:bpe} of the Appendix.
\section{Parity-aware Byte Pair Encoding}

Classical BPE chooses merges that maximize a \emph{global} frequency objective, implicitly favoring the compression of languages with a larger presence in the training corpus. Here we introduce \defn{Parity‑aware BPE}, which replaces this global objective with a \emph{max–min} criterion: at every step, it selects the merge that most improves the worst-compressed language.

\subsection{Greedy Fair-Max Objective}
Our adjustment to the Classical BPE objective (\Cref{eq:bpe_objective}) explicitly encodes our definition of tokenizer fairness: equality of compression rates across languages. 
Formally, Parity-aware BPE seeks a merge list $\merges=[\merge_{1},\dots,\merge_{K}]$ that maximizes the \emph{minimum} compression rate across languages:
\vspace{-2mm}
\begin{equation}\label{eq:parity_bpe_objective}
    \merges^\star = \operatorname*{argmax}_{\merges: |\merges| = K}\,\min_{\lang}\,\comprate(\lang; \tok_\merges) 
\end{equation}
\noindent 
We call this a fair-max objective; it trades a small amount of global compression for cross-lingual parity. 

\subsection{Algorithm}\label{sec:algorithm}
Parity-aware BPE retains the greedy iterative framework of Classical BPE but changes \emph{which} statistics drive each merge decision. 
At step $k$, it identifies the worst-compressed language under the current tokenizer, \ie, under the tokenizer defined by the merge list thus far ($\tok_{\merges_{<k}})$:
\vspace{-2mm}
\begin{equation}\label{eq:chosen_lang}
    \lang^{\star}
            \;=\; \arg\min_{\lang\in\languages}\comprate(\lang; \tok_{\merges_{<k}})
\end{equation}
It then uses Classical BPE's maximum pair-count merge selection criterion, but restricted to  $\corpus_{\lang^{\star}}$---the portion of the corpus in language $\lang^{\star}$.  
The rest of the algorithm follows the Classical BPE procedure: the chosen merge is applied to all texts (\ie, across $\corpus_\lang$ $\forall\lang$)\footnote{Crucially, this is what distinguishes our algorithm from a combination of monolingual merge lists \citep{DBLP:conf/nips/PetrovMTB23}, allowing us to find more ``compressive'' merges.} and the procedure is repeated for $k=1,\dots,K$, yielding the final merge list $\merges$. We providepseudocode in \Cref{alg:pabpe} of the Appendix.

\paragraph{Cross-lingual Compression Rate Comparison.}
Parity-aware BPE relies on the comparison of  $\comprate(\lang;\tok_\merges)$ across different $\lang$. 
The choice of normalization unit $u$ has a large impact on the measured $\comprate(\lang;\tok_\merges)$ and even when $u$ is held constant across measurements for different languages, if not considered carefully, the choice can introduce bias into the comparison. 
As concrete examples, certain normalization units are more appropriate in some languages than in others, \eg, whitespace‑delimited ``words'' are ill‑defined in many languages; although principled and universal, even normalizing by number of bytes can skew perceived compression because scripts differ greatly in average number of bytes per character (\eg, ASCII vs. UTF‑8 CJK). 
Parallel corpora provide a principled solution: computing compression rates over aligned texts (sentences, lines, or documents) normalizes by content, making cross‑language comparisons more meaningful. 
We therefore recommend the use of a parallel corpus for computing \Cref{eq:chosen_lang}.
Notably, this evaluation corpus need not be the same one used for computing subword pair frequency statistics, for which a larger corpus with only language annotation is necessary. 
For generality, we thus differentiate between the corpora used to compute frequency statistics and in computing \Cref{eq:chosen_lang}, referring to them as our training and development datasets, respectively. 
\Cref{alg:pabpe} makes this difference explicit. 
We present experimental results both with a separate, parallel development set and using a single (not parallel) multilingual dataset for all computations.

\paragraph{Complexity and Data Requirements.}
Relative to Classical BPE, Parity-aware BPE incurs only a $O(|\languages|)$ overhead per‑merge from recomputing the language‑specific compression rates on the dev set. 
Parity-aware BPE retains the same asymptotic complexity as Classical BPE, requiring only some modest additional bookkeeping.  
The need for a multi-parallel corpus can at first seem prohibitive, but several pragmatic design choices can reduce the burden of this requirement. 
A small multi-parallel dataset suffices to drive the max-min decision in \Cref{eq:chosen_lang}, and the training dataset need not be parallel. In addition, automatic language ID tools or script heuristics can help provide language labels when none are readily available. Also note that only the BPE learning phase differs; there is no algorithmic change to the tokenization function itself.

\subsection{Algorithmic Variants}

We now introduce several variants of Parity-aware BPE; these variants are designed to broaden its applicability to different developer goals and resource availability, as well as to address challenges identified in preliminary experiments.

\paragraph{Hybrid Parity-aware BPE.}
Developers may wish to have more fine-grained control of the balance between global compression and cross-lingual token count parity.  
We support these goals with a hybrid learning algorithm that uses the global objective of Classical BPE (\Cref{eq:bpe_objective}) for the first $J$ merges, then switches to the Parity-aware objective (\Cref{eq:parity_bpe_objective}) for another $K$ merges. $J$ and $K$ can be chosen by model developers to trade off global compression and fairness according to their priorities.

\paragraph{Moving-window Balancing.} There may be a point where the compression in a language no longer or barely improves, even if it is repeatedly chosen for the next merge. This could happen for several reasons: the development dataset (the dataset used to choose the language) may be too small or not match the domain or language variant of the training dataset;\footnote{\citet{kreutzer-etal-2022-quality} discuss possible quality issues such as wrong or ambiguous language codes.} alternatively, $\lvert\tok(\bytesequence)\rvert$ may approach the length of the pre-tokenized sequence, meaning there are no more possible merges. To prevent our algorithm from being ``stuck'' selecting the same language exclusively, we track the $W$ most recent languages selected in \Cref{eq:chosen_lang}, and do not select a language if it occurs more than $\alpha\frac{W}{|\languages|}$ times in this moving window.

\paragraph{Ratio-normalized (dev set-free) Language Selection Criterion.} When multi-parallel development data is partially or completely unavailable, or when the concept of a parallel corpus is not applicable for one of the data domains (\eg, for code or math data), developers can instead specify per-language compression targets explicitly. At a given merge step, the language whose current compression rate is furthest below its target is selected to determine the next merge.  This compression rate can be computed directly on the training set, bypassing the need for a development set. 

Formally, we replace the language selection criterion in \Cref{eq:chosen_lang} with
\begin{equation}
    \lang^\star = \arg\min_{\lang \in \languages} \; \frac{\comprate(\lang;\, \tok_{\merges_{<k}})}{r_\lang}
\end{equation}
where $r_\lang > 0$
is a user-specified target value for language $\lang$. 
    These ratios $r_\lang$ can be interpreted as desired \textbf{relative compression rates} across languages. For instance, setting all $r_\lang$ equal recovers an equal-compression objective (albeit measured on non-parallel data), while setting $r_\lang$ proportional to the average bytes per line in a reference parallel corpus approximates the content-based normalization of the standard dev-set approach without requiring that corpus at training time. Because of the baseline differences in per-language UTF-8 encoding sizes, we recommend the latter approach. Developers can also use heuristics to approximate suitable ratios.
\section{Experimental Setup}
We conduct experiments to evaluate the effectiveness of Parity-aware BPE, comparing it against Classical BPE. All tokenizers are byte-level.

\subsection{Tokenizer Training}\label{subsec:lang_selection}
\textbf{Training Data.}
We train two sets of tokenizers: (1) using the multilingual C4 (mC4) corpus~\citep{xue-etal-2021-mt5, 2019t5}\footnote{\url{https://huggingface.co/datasets/allenai/c4}} (2) using the \finewebtwo{} dataset \citep{penedo2025fineweb}. We focus on the results for tokenizers trained on \finewebtwo{}; mC4 results can be found in \cref{app:abblation_studies}.\\
\textbf{Development Data.}
We use the dev portion of FLORES+~\citep{nllb-24}\footnote{\url{https://huggingface.co/datasets/openlanguagedata/flores_plus}} as our multilingual development corpus, \ie, for choosing the focus language at each merge step when training Parity-aware BPE tokenizers (\Cref{eq:chosen_lang}).
We also use this dataset to estimate per-language byte-level compression rate targets for the \textit{ratio} system; we then compare this value to the per-language byte-level compression rate of the training corpus to choose the focus language. \\
\textbf{Data Distribution.}
To investigate how the number of languages, their linguistic diversity, and the variety of writing systems influence tokenizers, we consider two language sets: one with 30 languages (\thirtylangs{}) and another with 60 languages (\sixtylangs{}). 
For each set, we construct two versions of the training data:
(1) \textbf{\unbalanced training:} In this setting, the tokenizer is trained on the naturally occurring mixture of languages, where high-resource languages exert greater influence on merge decisions due to their larger presence in the corpus. We determine the number of training examples per language using temperature-based sampling ($\tau = 3.3$, following prior work~\citealp{raffel2020exploring, conneau-etal2020-unsupervised, xue2021mt5}). This setup reflects standard practice in multilingual LLM development, where the tokenizer training distribution mirrors the overall pretraining corpus distribution to avoid domain shift between tokenizer and model training~\citep{DBLP:journals/corr/abs-2402-01035, ali2024tokenizer}.
We derive the per-language sampling parameters according to per-language data proportions in the mC4 dataset and use the same values for creating both the mC4 and \finewebtwo-based training datasets. 
(2) \textbf{\balanced training:} Here, we sample the same number of documents per language. This ensures that each language contributes equally to the tokenizer's training, even without explicit interventions such as Parity-aware BPE.  
We present results for the \unbalanced datasets here, as this is arguably the more realistic setting, with results for the \balanced setting shown in \cref{app:abblation_studies}. 
To enable tokenizer analyses as a function of different dataset qualities, we categorize the languages in each set based on the amount of training data available and the script family, performing some of our analyses by these categories. 
Languages with \(>1\text{M}\) examples are considered high-resource; those with  \(500\text{k} - 1\text{M}\) examples are medium-resource; and those with fewer than \(<500\text{k}\) examples are classified as low-resource.
The full list of languages included in each set and the script family groupings are presented in \cref{tab:language_list} of the Appendix.\\
\textbf{Hyperparameter Settings.} 
We look at vocabulary sizes $128k$ and $256k$.
For \textit{hybrid} systems, we learn half of the merges using the global strategy, and the second half using the Parity-aware strategy (\ie, $J=K = \frac{1}{2}|\merges|)$. For systems with moving-window balancing (\textit{window}), we use a window size of 100, and $\alpha=2$. More details and ablations over these hyperparameters can be found in \cref{app:hyperparams} and \cref{app:abblation_studies}.

\subsection{Evaluation} 
Our evaluations consist of task‑independent tokenizer properties (intrinsic metrics) and downstream model performance (extrinsic metrics).
Within a set of evaluations, we fix the data distribution (\balanced or \unbalanced) and vocabulary size ($128k$ or $256k$).

\subsubsection{Intrinsic Metrics}\label{sec:intrinsic}
We measure a variety of intrinsic tokenizer metrics on the devtest portion of FLORES+.
These metrics encompass basic tokenization properties, information-theoretic measures, cross-linguistic fairness, and morphological alignment.  
Because of the parallel nature of FLORES+, we can use ``lines'' as our normalization unit for most metrics, which we motivate as a good default normalization unit in \Cref{sec:algorithm}. 
For some intrinsic metrics, though, other normalization units are more appropriate, \eg, words or characters may be more appropriate when morphologically motivated units can better reflect linguistic structure. 
We make this deviation from the default normalization unit explicit when applicable. 
For the sake of space, we provide brief metric descriptions here; more detailed metric definitions can be found in \cref{app:intrinsic}.
\begin{itemize}[leftmargin=*,noitemsep, topsep=2pt]
\item \defn{Fertility} measures the average number of tokens produced per normalization unit by a tokenizer; whitespace-delimited words are often the unit of interest (and are the units used in our computations). In this case, fertility quantifies how many tokens (on average) a word is broken up into~\citep{DBLP:conf/acl/RustPVRG20}.  
\item \defn{Compression Rate (CR)} (as defined in \cref{sec:tokenization_background}) is a measure of the degree to which a unit of text has been shrunk after applying the given tokenizer (higher is better). \looseness=-1 

\item \defn{Vocabulary Utilization} is the fraction of the tokenizer’s vocabulary that actually appears in the evaluation corpus. Low utilization for a language signals wasted capacity or, when there are large differences across languages, a vocabulary allocation that is biased towards certain languages.  

\item \defn{Gini Token Inequality Coefficient}  \citep{meister_tokenizer_analysis_2025} adapts the Gini coefficient to the per-language tokenization cost distribution (\eg, tokens per line (document) in a parallel corpus). Values near 0 mean equal cost across languages; values closer to 1 indicate inequality. 

\item \defn{MorphScore} \citep{arnett2025alignment} measures how well token boundaries align with true morpheme boundaries, computed as morpheme-level precision/recall (and F1). High scores mean tokens respect morphological structure; low precision implies over-segmentation, while low recall may suggest under-segmentation.
\end{itemize}
\noindent For completeness, we also track Type–Token Ratio and Average Token Rank \citep[vocabulary diversity;][]{limisiewicz-etal-2023-tokenization} as well as Rényi entropies \citep[distributional concentration;][]{zouhar-etal-2023-tokenization}, evaluated using the TokEval suite \citep{meister_tokenizer_analysis_2025}. 

\begin{table*}
\centering
\small
\adjustbox{max width=\linewidth}{

\begin{tabular}{lcccccccc}
\toprule
Tokenizer & \makecell{Comp. Rate ($\uparrow$)} & \makecell{TTR ($\uparrow$)} & \makecell{Vocab \\Utilization  ($\uparrow$)} & \makecell{Fertility \\ ($\downarrow$)} & \makecell{Rényi \\ ($\alpha$=2.5)  ($\uparrow$)} & \makecell{Gini ($\downarrow$)} & \makecell{MorphScore \\ Precision  ($\uparrow$)} & \makecell{MorphScore \\ Recall  ($\uparrow$)} \\
\midrule
BPE & 0.0275 & 0.0777 & 67.0\% & \textbf{3.990} & 0.49 & 0.064 & 0.537 & 0.659 \\
PA-BPE & 0.0276 & 0.0819 & 70.4\% & 4.080 & 0.49 & \textbf{0.007} & 0.539 & 0.671 \\
PA-BPE (window) & 0.0278 & \textbf{0.0838} & \textbf{71.6\%} & 4.042 & 0.48 & 0.009 & 0.546 & \textbf{0.677} \\
PA-BPE (hybrid) & 0.0278 & 0.0817 & 69.8\% & 4.056 & 0.49 & 0.015 & 0.541 & 0.667 \\
PA-BPE (hybrid+window) & \textbf{0.0279} & 0.0832 & 70.8\% & 4.029 & 0.48 & 0.019 & \textbf{0.546} & 0.670 \\
PA-BPE (ratios) & 0.0270 & 0.0737 & 64.9\% & 4.398 & 0.49 & 0.040 & 0.532 & 0.669 \\
\bottomrule
\end{tabular}

}\caption{Intrinsic evaluation of $128k$ tokenizers trained on the (\unbalanced) \thirtylangs{} dataset, evaluated on the corresponding \thirtylangs{} subset of the FLORES+ dataset. With the exception of MorphScore---which is macro-averaged across available languages---values are global statistics across the parallel corpus. }
\label{tab:full_unbalanced_30langs_128k}
\vspace{-5mm}
\end{table*}
\subsubsection{Extrinsic Metrics.}
For extrinsic evaluation, we train models using different tokenizers and assess their performance across a range of downstream tasks.
\paragraph{Model Architecture and Pretraining Data.}
We train decoder-only Transformer models~\cite{vaswani2017attention} following the LLaMA architecture~\cite{touvron2023llama} with  $3$B parameters. 
We use tokenizers trained on mC4. LMs are trained on $100$B tokens from \finewebtwo{}. To create the model training dataset, we adopt temperature sampling with $\tau=3.3$, following recommendations from prior work~\citep{2019t5, conneau-etal2020-unsupervised}. Full details on model configurations and training parameters are provided in \cref{appendix:training_setup}. 

\paragraph{Benchmarks.}
We evaluate the models using perplexity on a held-out validation set from the respective pretraining datasets. 
In addition, we evaluate their multilingual performance using twelve widely adopted benchmarks that collectively cover a broad range of tasks, including reading comprehension, commonsense reasoning, semantic similarity, and knowledge-based evaluation. These benchmarks are chosen to provide reliable and informative signals (results clearly above random chance) for small-scale models (1-3B parameters), while also supporting zero-shot evaluation without requiring supervised finetuning. 
The selected benchmarks covering 22 languages are Belebele, mTruthfulQA, PAWS-X, XCodah, XCSQA, XNLI, XStoryCloze, XWinogrande, MMMLU, INCLUDE, Exams, and M3Exams.
Results are aggregated per language to produce a score for each model-language pair.
A full list of benchmarks and aggregation procedures is provided in \cref{appendix:benchmark_setup}.
\section{Results and Analysis}
\label{sec:results}
\begin{figure}\centering\includegraphics[width=\linewidth]{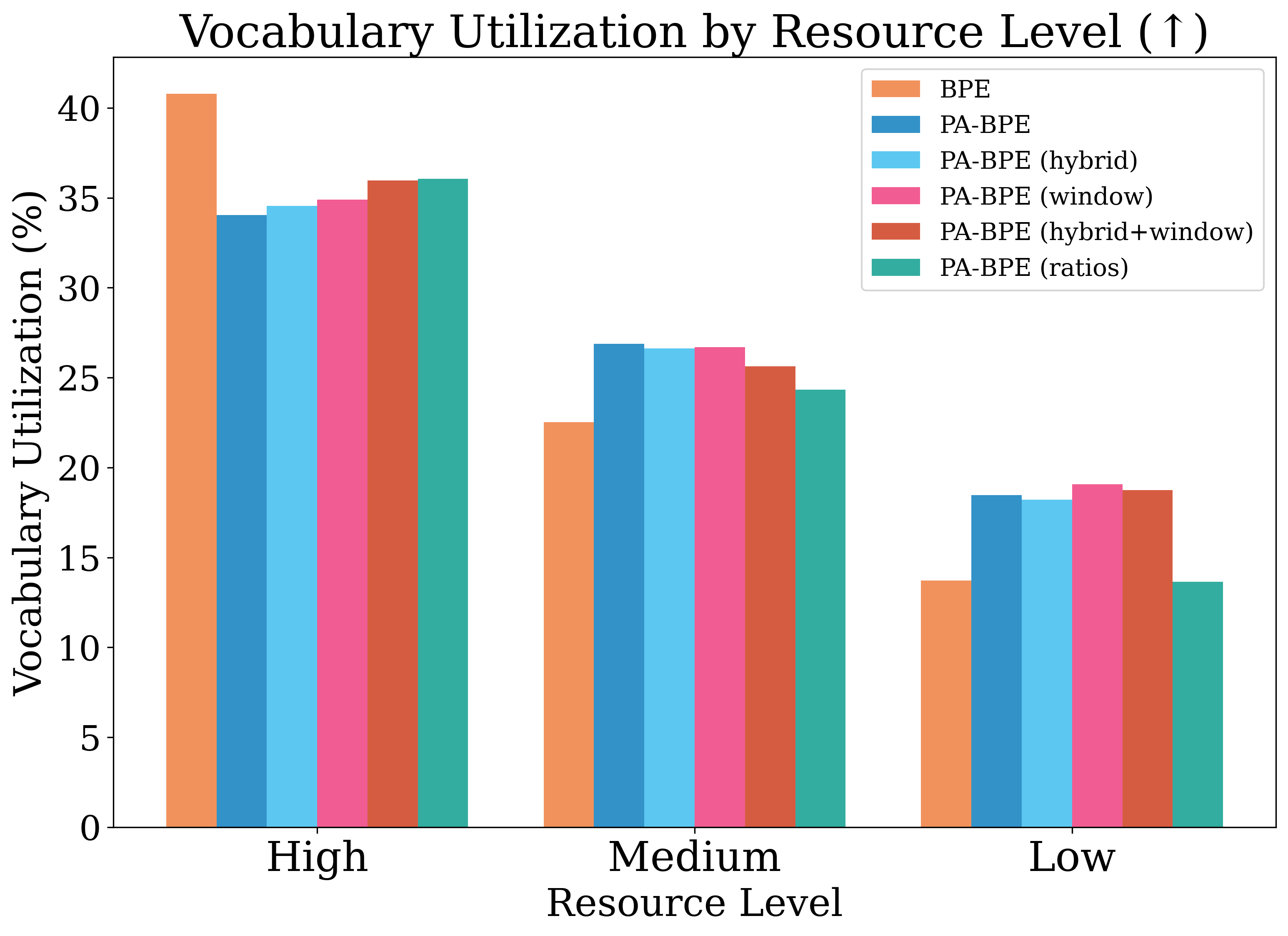}
    \caption{Vocabulary utilization for $128k$ tokenizers on the \unbalanced \thirtylangs{} dataset grouped by each language's resource level.\looseness=-1 }
    \label{fig:vocab_util_unbalanced_30langs_128k}
\end{figure}
\begin{figure*}
    \centering
    \includesvg[width=\linewidth]{figs/128k/30lang_unbalanced_fineweb2/faceted_plots/compression_rate_faceted}
    \caption{Per-language compression rate ($\uparrow$) of FLORES+ parallel sentences when tokenized using $128k$ tokenizers trained on the (\unbalanced) \thirtylangs{} per language.}
    \label{fig:comprate_per_lang_unbalanced_30langs_128k}
\end{figure*}

\subsection{Intrinsic Evaluation}

Results in \cref{tab:full_unbalanced_30langs_128k} show that all variants of Parity-aware BPE substantially reduce the Gini coefficient relative to Classical BPE (from $0.064$ to $\leq 0.019$ for dev-set variants), indicating more equitable token costs across languages. Among these variants, the base Parity-aware BPE achieves the lowest Gini, \ie, it is the ``fairest'' tokenizer. Average per-language vocabulary utilization also improves slightly for the dev-set variants compared to Classical BPE.
Global compression rates and Rényi entropies remain essentially identical across all BPE variants, confirming that the fairness gain does not come at a measurable cost to overall compression efficiency.  
Other intrinsic metrics remain relatively stable across all Parity-aware variants;
we treat these as auxiliary diagnostics, confirming that the fairness improvement does not negatively affect other tokenizer properties. 

We show per-language results for compression rate in \cref{fig:comprate_per_lang_unbalanced_30langs_128k}, which illustrates the core result for the method: Parity-aware tokenizers yield substantially more uniform compression across languages than Classical BPE, consistent with the Gini reduction reported in \cref{tab:full_unbalanced_30langs_128k}. 
Notably, as visible in both \cref{tab:full_unbalanced_30langs_128k} and \cref{fig:comprate_per_lang_unbalanced_30langs_128k}, the \emph{ratios} variant achieves the smallest Gini reduction of any of the Parity-aware variants. 
This is likely attributable to domain mismatch: the per-language compression targets are derived from bytes-per-line statistics in the parallel FLORES+ dataset, which is necessarily cross-lingually domain-aligned. However, the statistics used for merge selection are computed on \finewebtwo{}, whose domain and genre composition may differ substantially across languages.  
Because data from different domains can require different levels of tokenization granularity (\eg, technical terminology may demand finer-grained segmentation), FLORES-derived ratios may not be ideal for achieving content-controlled parity--—as operationalized by the Gini coefficient on FLORES+---on the \finewebtwo{} training corpus. Nevertheless, this variant remains practical alternative to improve upon Classical BPE when multi-parallel data is limited.

\begin{table}[ht]
\centering
\small
\resizebox{\columnwidth}{!}{

\begin{tabular}{l ccc c}
\toprule

Language & Classical BPE & \makecell{Parity-aware \\ (hybrid)} & \makecell{Parity-aware \\ (window+hybrid)} & \!Random  \\

\midrule
Arabic & 38.19 \scriptsize  $\pm$ 2.90 & \textbf{39.04} \scriptsize  $\pm$ 2.89 & 38.84 \scriptsize  $\pm$ 2.90 & 32.00 \\
Bengali & 24.95 \scriptsize  $\pm$ 3.09 & 23.54 \scriptsize  $\pm$ 2.98 & 23.91 \scriptsize  $\pm$ 3.01 & \textbf{25.00} \\
German & 32.92 \scriptsize  $\pm$ 3.14 & 34.78 \scriptsize  $\pm$ 3.66 & \textbf{36.82} \scriptsize  $\pm$ 4.04 & 30.62 \\
Greek & 41.95 \scriptsize  $\pm$ 3.18 & 42.55 \scriptsize  $\pm$ 3.22 & \textbf{43.16} \scriptsize  $\pm$ 3.25 & 37.50 \\
Spanish & 37.53 \scriptsize  $\pm$ 2.66 & 38.83 \scriptsize  $\pm$ 2.71 & \textbf{39.30} \scriptsize  $\pm$ 2.75 & 32.77 \\
Persian & \textbf{42.80} \scriptsize  $\pm$ 5.39 & 39.15 \scriptsize  $\pm$ 5.27 & 39.15 \scriptsize  $\pm$ 5.27 & 25.00 \\
French & \textbf{38.67} \scriptsize  $\pm$ 3.90 & 36.59 \scriptsize  $\pm$ 2.84 & 37.10 \scriptsize  $\pm$ 2.82 & 32.00 \\
Hindi & \textbf{33.92} \scriptsize  $\pm$ 2.25 & 33.92 \scriptsize  $\pm$ 2.24 & 33.86 \scriptsize  $\pm$ 2.24 & 30.62 \\
Indonesian & 38.95 \scriptsize  $\pm$ 2.62 & \textbf{40.55} \scriptsize  $\pm$ 2.66 & 40.46 \scriptsize  $\pm$ 2.66 & 35.00 \\
Italian & 32.82 \scriptsize  $\pm$ 2.86 & \textbf{35.01} \scriptsize  $\pm$ 3.00 & 34.62 \scriptsize  $\pm$ 2.98 & 27.22 \\
Japanese & 37.43 \scriptsize  $\pm$ 2.39 & \textbf{37.45} \scriptsize  $\pm$ 2.39 & 37.43 \scriptsize  $\pm$ 2.39 & 34.00 \\
Korean & 33.00 \scriptsize  $\pm$ 5.22 & 33.00 \scriptsize  $\pm$ 5.22 & \textbf{34.33} \scriptsize  $\pm$ 5.29 & 25.00 \\
Polish & 29.75 \scriptsize  $\pm$ 2.50 & \textbf{31.14} \scriptsize  $\pm$ 2.60 & 28.97 \scriptsize  $\pm$ 2.49 & 23.75 \\
Portuguese & \textbf{33.63} \scriptsize  $\pm$ 2.81 & 33.15 \scriptsize  $\pm$ 2.77 & 33.06 \scriptsize  $\pm$ 2.77 & 27.50 \\
Russian & 36.36 \scriptsize  $\pm$ 2.27 & 36.21 \scriptsize  $\pm$ 2.26 & \textbf{36.57} \scriptsize  $\pm$ 2.28 & 32.77 \\
Tamil & 31.32 \scriptsize  $\pm$ 2.81 & \textbf{32.25} \scriptsize  $\pm$ 2.90 & 32.19 \scriptsize  $\pm$ 2.90 & 31.25 \\
Telugu & 32.73 \scriptsize  $\pm$ 2.61 & \textbf{33.52} \scriptsize  $\pm$ 2.61 & 33.26 \scriptsize  $\pm$ 2.61 & 30.00 \\
Turkish & \textbf{39.04} \scriptsize  $\pm$ 2.89 & 38.46 \scriptsize  $\pm$ 2.83 & 37.89 \scriptsize  $\pm$ 2.75 & 35.00 \\
Vietnamese & 33.69 \scriptsize  $\pm$ 2.31 & \textbf{33.87} \scriptsize  $\pm$ 2.27 & 33.80 \scriptsize  $\pm$ 2.30 & 29.50 \\
Chinese & 38.43 \scriptsize  $\pm$ 2.11 & \textbf{38.58} \scriptsize  $\pm$ 2.11 & 38.32 \scriptsize  $\pm$ 2.10 & 35.00 \\
English & 43.04 \scriptsize  $\pm$ 1.84 & \textbf{44.15} \scriptsize  $\pm$ 1.85 & 43.74 \scriptsize  $\pm$ 1.85 & 35.50 \\
Thai & 40.76 \scriptsize  $\pm$ 1.62 & 40.96 \scriptsize  $\pm$ 1.63 & \textbf{41.06} \scriptsize  $\pm$ 1.63 & 37.50 \\
\bottomrule
\end{tabular}
}
\caption{Average downstream performance (accuracy \%) across 12 multilingual benchmarks (tokenizers trained on the (\unbalanced) \thirtylangs{} dataset with $128k$ vocab size). 
The \textbf{Random} column shows the expected accuracy of a random classifier. 
Best performance per language is bolded. Benchmark details for each language are provided in \cref{appendix:benchmark_setup,tab:language_benchmark-coverage}, respectively.\looseness=-1}
\vspace{-1mm}
\label{tab:benchmark_results}
\end{table}

\cref{fig:vocab_util_unbalanced_30langs_128k} presents vocabulary utilization grouped by resource tier, as defined in \cref{subsec:lang_selection}.
Parity-aware tokenizers provide more consistent usage across languages, evening out the vocabulary allocation to high vs.\ low resource languages in comparison to Classical BPE. As with other intrinsic metrics, the ratios variant
exhibits the least consistency in cross-resource-level vocabulary utilization among the Parity-aware variants.

\paragraph{Ablations.} 
We report additional experiments in \cref{app:abblation_studies} and \cref{app:hyperparams}, varying the tokenizer training corpus (mC4), vocabulary size (256k), per-language data allocation (balanced vs.\ unbalanced), window size for the window variant ($W=50,100, 150, 200$), dev set size ($N=50,100, 300, 1000$), and number of languages (30 vs.\ 60). We see consistency in results across all settings: Parity-aware BPE variants achieve substantially lower Gini coefficients than Classical BPE, while compression rates and other intrinsic metrics remain comparable.

\subsection{Extrinsic Evaluation}

\cref{tab:benchmark_results} presents the performance of LMs trained with three different $128k$ tokenizers: Classical BPE, Hybrid Parity-aware BPE, and Hybrid Parity-aware BPE (moving-window), evaluated on a subset of the \thirtylangs{} set (22 languages).
For each language, we report mean performance, standard errors, and a random baseline to account for varying benchmark counts (\cref{tab:language_benchmark-coverage}).
We treat downstream evaluation as a regression check to assess whether the fairness gains come at the expense of downstream performance.
Results indicate that \pbpe{} maintains performance across languages:
for models trained with the hybrid variant, we see nominal gains in 14 languages and nominal declines in 6. Individual differences are generally within standard errors, and thus, we conclude that we find no evidence that Parity-aware tokenizers would compromise downstream LM performance. 
Additionally, per-language perplexity in Appendix  \cref{fig:per-lang-perp} show that Classical BPE produces models for which a subset of languages have notably higher perplexity. 
These results suggest that models trained with Parity-aware tokenizers can exhibit more uniform perplexity across languages.

\section{Related Work}

\paragraph{Multilingual Tokenization.}
Despite their popularity, BPE and related subword tokenization methods often underperform in multilingual settings due to limited handling of spelling variation and morphological complexity \citep{bostrom-durrett-2020-byte}.
A tokenizer's tokenization parity and fertility directly impact both computational cost and model performance \citep{lesci-etal-2025-causal}. Prior work has explored vocabulary allocation strategies: \citet{DBLP:conf/amta/ZhangCGCWBG22} show that larger vocabularies improve NMT robustness across scripts, while \citet{DBLP:conf/emnlp/GowdaM20} demonstrate that tuning BPE merges can mitigate sequence length issues. \citet{DBLP:conf/acl/RustPVRG20} find that integrating specialized monolingual tokenizers into multilingual systems can boost performance; however, recent evidence suggests that optimal vocabulary size depends on the task and model \citep{DBLP:journals/corr/abs-2402-01035}.
For multilingual vocabulary construction, \citet{Chung2020} explore clustering-based sharing of subwords across languages, and \citet{limisiewicz-etal-2023-tokenization} propose an explicit tokenizer-merging algorithm to combine per-language vocabularies.
Finally, tokenization-free models such as \mbox{\textsc{canine}} \citep{DBLP:journals/tacl/ClarkGTW22} and ByT5 \citep{DBLP:journals/tacl/XueBCANKRR22} offer an alternative approach for improved multilingual handling.

\paragraph{Tokenization Bias and Recent Advances.} 
Recent research highlights biases from tokenization in LLMs.
While \citet{wan2022fairness} argue that character‑ and byte‑level representations are intrinsically fair,\footnote{They report more random performance with these representations, which we do not view as ``fairer.''} other studies \citep{DBLP:conf/nips/PetrovMTB23,DBLP:conf/emnlp/Ahia0GKMST23} show that tokenization disparities across languages; even at character and byte levels; affect costs, latency, and contextual understanding. This has spurred efforts like Aya~\cite{aryabumi2024aya} and methods to mitigate tokenization unfairness \citep{DBLP:journals/corr/abs-2404-17790,DBLP:conf/coling/AbboudO24,limisiewicz2024myte}. Although newer character- and byte-level models use compression techniques such as entropy-based patching~\cite{DBLP:conf/acl/PagnoniP0NMLZYW25}, cross-lingual parity of these representations remains unstudied. 
\section{Discussion and Conclusion}

Tokenizers optimized using standard algorithms can lead to disparities in users' costs and experiences depending on language choice. 
Parity‑optimized tokenization can remedy this by explicitly balancing compression across languages, enabling fairer treatment of users of low‑resource languages.
Parity‑aware BPE operationalizes this idea; it is designed to improve cross‑lingual tokenization parity. Our experiments confirm its effectiveness:
we observe up to an $89\%$ reduction in the Gini token inequality coefficient in comparison to Classical BPE, while compression rates across all BPE variants remain very close. 
Crucially, this fairness gain does not come at the expense of downstream quality: across 22 languages, accuracy differences between Parity-aware and Classical BPE tokenizers fall within standard errors for nearly all languages (\cref{tab:benchmark_results}). In other words, we find no evidence of systematic downstream degradation from using Parity-aware BPE.

Overall, the trade-offs associated for using Parity-aware BPE are minimal. From the model‑developer’s perspective, Parity-aware BPE is a drop‑in replacement: it requires no architectural changes and only minimal modifications to the tokenizer training pipeline. Concretely, inference remains identical to that Classical BPE. During the learning stage, the algorithm adds only an $\mathcal{O}(\languages)$ pass per merge to recompute language‑level compression rates on a dev corpus, leaving the asymptotic complexity identical to Classical BPE. Moreover, we find empirically that a small, sentence‑aligned development set is sufficient to drive the fair‑max decision (\cref{sec:dev-set-size}). When resource or domain mismatches make full equality undesirable, hybrid, moving‑window, and ratio-normalized variants further let practitioners trade off global compression versus strict parity; our empirical results validate that these variants perform well in practice. 

Improving the equity of the tokenization step in the NLP pipeline is therefore not just desirable but feasible. 
With a simple modification---selecting merges that benefit the worst-compressed language---Parity‑aware BPE substantially reduces the hidden ``token tax'' imposed on speakers of low‑resource languages without sacrificing compression or task accuracy. 

Future research can extend this agenda by adapting the ``fair-max'' objective to alternative tokenization schemes and diverse modalities, such as speech and vision. For merge-based methods like WordPiece, selection criteria can be modified to prioritize merges that provide the greatest likelihood gains for the most poorly compressed language. Similarly, for non-merge-based algorithms like UnigramLM, the parity objective can be integrated into the pruning stage to preserve tokens essential for low-resource language efficiency. Moreover, future work should develop robust benchmarks and metrics for fairness assessments that go beyond simple compression parity to ensure truly equitable processing across all linguistic and sensory inputs.
\section*{Limitations}

Several sources of algorithmic bias can cause cross-lingual inequity in token counts to persist even when using Parity-aware BPE.  First, the standard variants of Parity-aware BPE  use parallel corpora to estimate per‑language costs; in domains where aligned documents are unavailable or the per-languages ratios in the ratio variant can only be heuristically estimated, the algorithm's ability to identify the truly worst-compressed language at each merge step may be degraded.   
Second, BPE cannot compress a text beyond its pre-tokenized length. If $\lvert\tok(\bytesequence)\rvert$ approaches the length of the pre-tokenized sequence for one language, any BPE variant (including Parity-aware BPE) will yield diminishing compression gains for that language. In particular, when using the moving-window balancing variant of Parity-aware BPE, the algorithm responds by relaxing the parity objective. This will also allow some cross-lingual disparities to persist. 

Future work has several opportunities to address these limitations: better approaches to estimating ratios for the ratio variant, alternative variants that bypass the need for a parallel development set entirely, and alternative pretokenization strategies that ensure some languages are not disadvantaged by the pretoken splitting boundaries---for example, via integration with an approach such as SuperBPE \cite{liu2026superbpe}.

More broadly, while we consider 60 languages and two vocabulary sizes, the interplay between tokenization parity and model scaling still needs to be explored for much larger models, larger language sets and for code or multimodal inputs.  
Finally, fairness in this work is defined purely in terms of token counts. While we measure other potential quantifications of fairness (\eg, morphological alignment), there are still other notions that are unaccounted for. We leave optimization for these metrics during tokenizer learning to future work.
\section*{Acknowledgement}

RS acknowledges support by the Swiss National Science Foundation through the MUTAMUR project (no. 213976). 
We gratefully acknowledge the support of the Swiss National Science Foundation (No. 215390), Innosuisse (PFFS-21-29), the European Research Council (Starting grant no. 101222478, RESPECT-LM), the AI2050 program at Schmidt Sciences (Grant \#G-25-69783), Sony Group Corporation, and the Swiss National Supercomputing Center (CSCS) in the form of an infrastructure engineering and development project.
We thank Tiago Pimentel for his feedback on earlier drafts of this manuscript.

\bibliography{anthology,custom}

\newpage
\clearpage
\appendix
\section{Pseudocode}\label{app:pseudocode}

\begin{algorithm}[h]
\caption{Algorithm for learning $\merges$ using Classical BPE.}\label{alg:bpe}
\DontPrintSemicolon
\KwIn{Corpus $\corpus$; number of merges $K$}
\KwOut{Vocabulary $\mathcal{V}_K$; merge sequence $\merges_K$}
\BlankLine

$\mathcal{V}_0 \leftarrow \mathcal{B}$\;
$\merges_0 \leftarrow \langle\,\rangle$\;
\BlankLine
\For{$k \gets 1$ \KwTo $K$}{
    \tcp{\small Count all adjacent token pairs}
    $\textit{Pairs} \leftarrow \{\}$\;
    \ForEach{occurrence of consecutive tokens $v\, v'$ in $\corpus$ where $v, v' \in \vocab_{k-1}$}{
        $\textit{Pairs}[(v,v')] \leftarrow \textit{Pairs}[(v,v')] + 1$\;
    }
    \BlankLine
    
    $(v^\star, v'^\star) \gets \arg\max_{(v,v')} \textit{Pairs}[(v,v')]$\;
    
    $w^\star \gets v^\star \!\circ\! v'^\star$\;
    
    \tcp{\small Update vocabulary and merge sequence}
    $\mathcal{V}_k \leftarrow \mathcal{V}_{k-1} \cup \{w^\star\}$\;
    $\merges_k \leftarrow \merges_{k-1}  + \langle (v^\star, v'^\star)\rangle$\;
    
    \tcp{\small Replace all occurrences in corpus}
    \ForEach{occurrence of $v^\star \, v'^\star$ in $\corpus$}{
        Replace $v^\star\, v'^\star$ with $w^\star$\;
    }
}

\Return{$\mathcal{V}_K$, $\merges_K$}
\BlankLine
\end{algorithm}

\begin{algorithm}[h!]
\caption{Algorithm for learning $\merges$ using Parity-aware Byte Pair Encoding with separate training and development sets.}
\label{alg:pabpe}
\DontPrintSemicolon
\KwIn{\small $\bigl\{\corpus_\lang\bigr\}_{\lang\in\languages}$ (multilingual training corpus);
\\  \,\,\,\,$\bigl\{\devcorpus_\lang\bigr\}_{\lang\in\languages}$ (multilingual development corpus);
\\ \,\,\,\,$K$ (number of merges) }
\BlankLine
\KwOut{\small $\vocab_K$ (vocabulary); $\merges_K$ (merge list) }
\BlankLine
$\vocab_0 \leftarrow \mathcal{B}$; \,\, $\merges_0 \leftarrow \langle\,\rangle$\;
\BlankLine
\For{$k \gets 1$ \KwTo $K$}{
    \tcp{\small Calculate compression rate for each language}
    \ForEach{language $\lang \in \languages$}{
        $\!\! \comprate(\devcorpus_\lang, \tok_{\merges_{<k}}) \gets
        \frac{\sum_{\bytesequence\in\devcorpus_\lang} |\bytesequence|_u}{\sum_{\bytesequence\in\devcorpus_\lang} |\tok_{\merges_{<k}}(\bytesequence)|}$\;
    }
    \BlankLine
    $\lang^\star \gets \arg\min_{\lang\in\languages} \,\,\comprate(\devcorpus_\lang, \tok_{\merges_{<k}})$\;
    \BlankLine
    
    \tcp{\small Consider token pairs only in $\corpus_{\lang^\star}$}
    $\textit{Pairs} \leftarrow \{\}$\;
    \ForEach{occurrence of consecutive tokens $v\,v'$ in $\mathcal{D}_{\lang^\star}$ where $v, v' \in \vocab_{k-1}$}{
        $\textit{Pairs}[(v,v')] \leftarrow \textit{Pairs}[(v,v')] + 1$\;
    }
    \BlankLine
        $(v^\star, v'^\star) \gets \arg\max_{(v,v')} \textit{Pairs}[(v,v')]$\;
    $w^\star \gets v^\star \!\circ\! v'^\star$\;
    \BlankLine
    
    \tcp{\small Update vocabulary and merge list}
    $\vocab_k \leftarrow \vocab_{k-1} \cup \{w^\star\}$\;
    $\merges_k \leftarrow \merges_{<k} + \langle (v^\star, v'^\star)\rangle$\;
    \BlankLine
    
    \tcp{\small Apply merge across all languages}
    
    \ForEach{language $\lang \in \languages$}{
        \ForEach{occurrence of $v^\star \, v'^\star$ in $\corpus_\lang$ and $\devcorpus_\lang$}{
            Replace $v^\star\, v'^\star$ with $w^\star$\;
        }
    }
}
\BlankLine

\Return{$\vocab_K$, $\merges_K$}
\end{algorithm}

\section{Intrinsic Tokenizer Evaluation Metrics}
\label{app:intrinsic}
We provide detailed descriptions of the intrinsic tokenizer metrics used in \cref{sec:results}, grouped by the general tokenizer characteristic the metric aims to assess. 
Metric formulae are defined in terms of our definition of a tokenizer $\tokenizer=(\vocab,\tok, \detok)$ given in \cref{sec:tokenization_background}. For this tokenizer,  we denote the empirical unigram frequency distribution of tokens $\token\in\vocab$ as $X_\tokenizer$, which is computed on our evaluation corpus.
\subsection{Vocabulary Usage}
  
  \paragraph{Vocabulary Utilization and Type-Token Ratio.}
  Vocabulary utilization measures the proportion of a tokenizer's full vocabulary that is actively used when processing a given corpus. For tokenizer $\tokenizer$ on corpus $\corpus$, we compute it as:
  \begin{equation}
  \text{VocabUtil}(\tokenizer) = \frac{|\{\token : \token \in \tok(\bytesequence), \bytesequence \in \corpus\}|}{|\vocab|}
  \end{equation}
  Here, the numerator counts the number of distinct tokens observed across
  the tokenization of all strings in the corpus.
   The type-token ratio quantifies lexical diversity by measuring the proportion of unique tokens (types) relative to the total number of tokens produced by
   a tokenizer:
  \begin{equation}
  \text{TTR}(\tokenizer) = \frac{|\{\token : \token \in \tok(\bytesequence), \bytesequence \in \corpus\}|}{\sum_{\bytesequence\in\corpus}|\tok(\bytesequence)|}
  \end{equation}
  where $|\tok(\bytesequence)|$ is the number of tokens produced by tokenizer $\tokenizer$ for input $\bytesequence$. In words, the numerator counts distinct token types and the denominator counts total tokens across the corpus.
  
High vocabulary utilization and type-token ratio indicate efficient use of the learned vocabulary; low values of these metrics for a particular language may suggest tokenizer bias, as only a small portion of the tokenizer's vocabulary is used/applicable for that language.

\paragraph{Average Token Rank.}
  Average token rank \citep{limisiewicz-etal-2023-tokenization} measures the typical position of tokens in a tokenized text within the frequency-ordered vocabulary. In more detail,  we compute the rank of each token (denoted as $\text{rank}(\token)$) in our unigram frequency distribution $X_\tokenizer$; rank 1 corresponds to the most frequent
  token. We compute average token rank across tokens in the evaluation corpus as:
  \begin{equation}
  \text{AvgRank}(\tokenizer) = \frac{\sum_{\bytesequence\in\corpus} \sum_{ \token \in \tok(\bytesequence)}
  \text{rank}(\token)}{\sum_{\bytesequence\in\corpus}|\tok(\bytesequence)|}
  \end{equation}
  This metric can be seen as another measure of the proportion of the vocabulary used by a tokenizer. Lower average ranks indicate that the tokenizer predominantly uses a small set of tokens, while higher averages suggest more diverse token usage, including rare vocabulary items. 
  When computed per language (\ie, when ranks are computed using the language's respective frequency distribution), systematic differences in average token rank across languages reveal vocabulary allocation bias.

\subsection{Information-theoretic Metrics}
  \paragraph{Compression Rate.}
We evaluate compression rate---as defined in \cref{eq:compression}---across a parallel corpus. As discussed in \cref{sec:algorithm}, this enables us to use lines (documents) as our normalization unit.
Recall that higher compression rates are generally desirable for computational efficiency in downstream tasks. In multilingual corpora, compression ratio disparities across languages indicate systematic tokenizer bias, where certain languages achieve better compression efficiency than others, potentially leading to unequal computational costs.

  \paragraph{Rényi Entropy.}
  We compute Rényi entropy of order $\alpha$ over the empirical unigram frequency distribution $X_\tokenizer$ for a given tokenizer $\tokenizer$ to capture different aspects of token distribution:
  \begin{equation}
  H_\alpha(X_\tokenizer) = \frac{1}{1-\alpha} \log_2 \left( \sum_{\token \in \vocab} p(\token)^\alpha \right)
  \end{equation}
  for $0 \alpha < \infty$; at $ \alpha =1$, the definition of Shannon entropy is typically adopted. Rényi entropy provides a parametric family of measures that emphasize different aspects of the distribution: $H_1$
  (Shannon entropy), $H_2$ (collision entropy), and $H_\infty$ (min-entropy). Rényi efficiency is Rényi entropy normalized by the size of the support, which is helpful for comparing tokenizers with different vocabulary sizes \citep{zouhar-etal-2023-tokenization}. As all of our comparisons are between tokenizers of the same vocabulary size, we omit this normalization step and compare entropies directly. 
  
\subsection{Morphological and Multilingual Fairness Metrics}
  \paragraph{Fertility.}
  Fertility measures the average number of tokens produced per unit (word, character, or byte) by a tokenizer; the unit of interest for fertility is often the \emph{word}, in which case, fertility quantifies how many tokens (on average) a word is broken up into. We use words as our normalization unit in our computations, as determined by the HuggingFace Whitespace Pretokenizer. 
  We formally define tokenizer fertility for a given corpus $\corpus$ as:
  \begin{equation}
  \text{Fertility}(\tokenizer) = \frac{\sum_{\bytesequence\in\corpus}|\tok(\bytesequence)|}{\sum_{\bytesequence\in\corpus}|\bytesequence|_u}
  \end{equation}
  This metric can give a sense for the computational efficiency imbued by a tokenizer, as well as for sequence length estimates for downstream modeling tasks.

  \paragraph{MorphScore.}
 MorphScore \citep{arnett2025alignment}
evaluates tokenizer quality through morpheme-level precision and recall, measuring how well tokenizers preserve morphological information during segmentation. We point the reader to the original work.
Differences in cross-language MorphScore reveal how consistently a tokenizer's sub‑token boundaries align with true morpheme boundaries. A higher score in one language than another indicates that the tokenizer preserves that language's morphological structure more faithfully. MorphScore provides a notion of both precision and recall (we point the reader to the original work for the exact description of the computation). Low precision indicates tokenizer oversegmentation; low recall is suggestive of \emph{under} segmentation.

\paragraph{Gini Token Inequality Coefficient.}
  We use an adaptation of the Gini coefficient---often used as a measure of economic inequality---to encapsulate tokenizer fairness across languages \citep{meister_tokenizer_analysis_2025}. 
  Formally, let $c_1 \leq c_2 \leq \ldots \leq c_n$ be the ``costs'' under a given tokenizer $\tokenizer$ for languages $\mathcal{L} = \{l_1, l_2, \ldots, l_n\}$. Here, we quantify cost as the average number of tokens it takes to encode the unit of interest (\eg, a byte, word or line);\footnote{This is equivalent to fertility, or the inverse of the compression rate.} when using a parallel corpus, this can be cost per line (document), which controls for discrepancies between average character byte lengths across different scripts. 
   The Gini coefficient for tokenizer $\tokenizer$ is then:
  \begin{equation}
  \text{Gini}(\tokenizer) = \frac{1}{n} \left( n + 1 - 2 \frac{\sum_{i=1}^n (n + 1 - i) c_i}{\sum_{i=1}^n c_i} \right)
  \end{equation}
  Values range from 0 (completely equal costs across languages) to 1 (maximum inequality). This metric condenses multilingual tokenizer fairness into a single number by measuring the degree of inequality in computational costs across languages; lower Gini coefficients indicate more equitable tokenizer compression across languages, while higher values suggest systematic bias toward certain languages.

\section{Hyperparameter Selection}
\label{app:hyperparams}

\subsection{Hybrid Parity-aware BPE}

For the hybrid Parity-aware tokenizer, we set the number of Parity-aware merges $K$ equal to the number of classical BPE merges $J$ (\ie, $K = J$, with each set to either 64$k$ or 128$k$). This choice represents a midpoint between classical BPE ($J > 0, K = 0$) and fully Parity-aware BPE ($J = 0, K > 0$). Alternative settings allow practitioners to trade off efficiency on high-resource languages (\eg, English) against cross-lingual fairness. Because no single operating point is objectively optimal across use cases, we do not claim a universally optimal choice for this trade-off.

\subsection{Moving-Window Balancing}

The moving-window balancing mechanism is designed to prevent degenerate cases in which a single language repeatedly consumes merge operations despite diminishing compression gains. The window size parameter ranges from 1 (enforcing uniform language selection across merges) to the number of languages $N$ (equivalent to no constraint). In a 30-language setting, this implies that even if compression improvements for a given language stagnate, it can consume at most approximately $1/15$ of merge operations. Increasing the window size causes the behavior to increasingly resemble the unbalanced setting.

Because this mechanism primarily guards against pathological merge allocation rather than directly optimizing compression, we do not expect the window size to be a sensitive hyperparameter.

To validate this intuition, we conduct ablation experiments varying the window size in the moving-window balancing variant. Table~\ref{tab:window_ablation} reports intrinsic tokenizer metrics for window sizes ranging from 50 to 200. Average compression rate, morphological plausibility, and other intrinsic metrics remain largely stable across window sizes. Empirically, the windowing mechanism prevents repeated selection of the same language caused by development--training set mismatch, without introducing sensitivity to the specific window size. These results suggest that the moving-window approach can be applied robustly without careful tuning.

\begin{table*}[t]
\centering
\small

\begin{tabular}{lcccccccc}
\toprule
Tokenizer & \makecell{Comp. Rate \\ ($\uparrow$)} & \makecell{TTR \\ ($\uparrow$)} & \makecell{Vocab \\ Utilization \\ ($\uparrow$)} & \makecell{Fertility \\ ($\downarrow$)} & \makecell{Rényi \\ ($\alpha$=2.5) \\ ($\uparrow$)} & \makecell{Gini \\ ($\downarrow$)} & \makecell{MorphScore \\ Precision \\ ($\uparrow$)} & \makecell{MorphScore \\ Recall \\ ($\uparrow$)} \\
\midrule
PA-BPE (W=50) & 0.0277 & 0.0832 & 71.3\% & 4.056 & \textbf{0.48} & \textbf{0.008} & 0.542 & 0.673 \\
PA-BPE (W=100) & \textbf{0.0278} & \textbf{0.0838} & \textbf{71.6\%} & \textbf{4.042} & 0.48 & 0.009 & \textbf{0.546} & \textbf{0.677} \\
PA-BPE (W=150) & 0.0277 & 0.0835 & 71.5\% & 4.049 & 0.48 & 0.009 & 0.543 & 0.674 \\
PA-BPE (W=200) & 0.0278 & 0.0837 & 71.6\% & 4.043 & 0.48 & 0.009 & 0.546 & 0.677 \\
\bottomrule
\end{tabular}

\caption{Intrinsic tokenizer evaluation metrics for ablations over the moving-window size. Results indicate minimal sensitivity to the window size parameter.}
\label{tab:window_ablation}
\end{table*}

\subsection{Parallel Development Set Size}\label{sec:dev-set-size}

We further study the effect of the size of the parallel development set used in cross-lingual compression rate comparisons for language selection (Eq.~7). Table~\ref{tab:devset_ablation} reports results for development set sizes of 100, 300, and 1000 examples.

The results indicate that relatively small development sets suffice to capture most of the fairness improvements. A parallel development set of 100 examples achieves a 67\% reduction in the Gini Inequality Coefficient relative to the classical baseline, compared to a 72\% reduction when using the full 1000-example development set. Global compression rates remain essentially unchanged across settings, suggesting that incorporating a parallel development set is feasible without imposing substantial data or computational burdens.

\begin{table*}[t]
\centering
\small

\begin{tabular}{lcccccccc}
\toprule
Tokenizer & \makecell{Comp. Rate \\ ($\uparrow$)} & \makecell{TTR \\ ($\uparrow$)} & \makecell{Vocab \\ Utilization \\ ($\uparrow$)} & \makecell{Fertility \\ ($\downarrow$)} & \makecell{Rényi \\ ($\alpha$=2.5) \\ ($\uparrow$)} & \makecell{Gini \\ ($\downarrow$)} & \makecell{MorphScore \\ Precision \\ ($\uparrow$)} & \makecell{MorphScore \\ Recall \\ ($\uparrow$)} \\
\midrule
PA-BPE (N=100) & 0.0277 & \textbf{0.0825} & \textbf{70.8\%} & 3.973 & 0.49 & 0.014 & 0.538 & 0.669 \\
PA-BPE (N=300) & 0.0277 & 0.0823 & 70.7\% & 4.019 & 0.49 & 0.012 & 0.539 & \textbf{0.672} \\
PA-BPE (N=997) & 0.0276 & 0.0819 & 70.4\% & 4.080 & \textbf{0.49} & \textbf{0.007} & 0.539 & 0.671 \\
PA-BPE hybrid (N=100) & \textbf{0.0278} & 0.0821 & 70.1\% & \textbf{3.967} & 0.48 & 0.022 & 0.540 & 0.666 \\
PA-BPE hybrid (N=300) & 0.0278 & 0.0819 & 70.0\% & 4.015 & 0.49 & 0.019 & 0.540 & 0.667 \\
PA-BPE hybrid (N=997) & 0.0278 & 0.0817 & 69.8\% & 4.056 & 0.49 & 0.015 & \textbf{0.541} & 0.667 \\
\bottomrule
\end{tabular}

\caption{Intrinsic tokenizer evaluation metrics for ablations over the size of the parallel development set. Smaller development sets retain most fairness gains. Other intrinsic metrics are very stable across different choices.}
\label{tab:devset_ablation}
\end{table*}

\section{Additional Results and Ablation Studies}

The following ablation studies are all with tokenizers trained on the mC4 dataset. We ablate tokenizer vocabulary size, number of languages in the training dataset and per-language dataset balance (\ie, whether dataset composition is \balanced across languages or \unbalanced). We also include a comparison against the UnigramLM tokenization algorithm \citep{DBLP:conf/acl/Kudo18}; we use the HuggingFace implementation of the tokenizer.

\begin{figure*}
    \centering
\includesvg[width=\linewidth]{figs/128k/30lang_unbalanced_mc4/faceted_plots/vocabulary_utilization_faceted}
    \caption{Vocabulary utilization of $128k$ tokenizers on the (\unbalanced) \thirtylangs{} per language. Dashed lines indicate the global average. }
    \label{fig:vocab_util_per_lang_unbalanced_30langs_128k}
\end{figure*}

\label{app:abblation_studies}
In this section, we present the results of our ablation studies.
\cref{tab:full_unbalanced_60langs_128k} reports the intrinsic evaluation of tokenizers with a $128k$ vocabulary size on the (\unbalanced) \sixtylangs{} dataset.
\cref{tab:full_balanced_30langs_128k} shows the corresponding results for the (\balanced) \thirtylangs{} dataset, also with a $128k$ vocabulary size.
Finally, \cref{tab:full_unbalanced_30langs_256k} presents the intrinsic evaluation of tokenizers with a $256k$ vocabulary size on the (\unbalanced) \thirtylangs{} dataset.
Together, these results demonstrate the effectiveness of Parity-aware BPE across different language settings, vocabulary sizes, and data distributions.

\begin{figure*}
    \centering
    \begin{subfigure}[t]{0.49\textwidth}
        \centering
        \includegraphics[width=\linewidth]{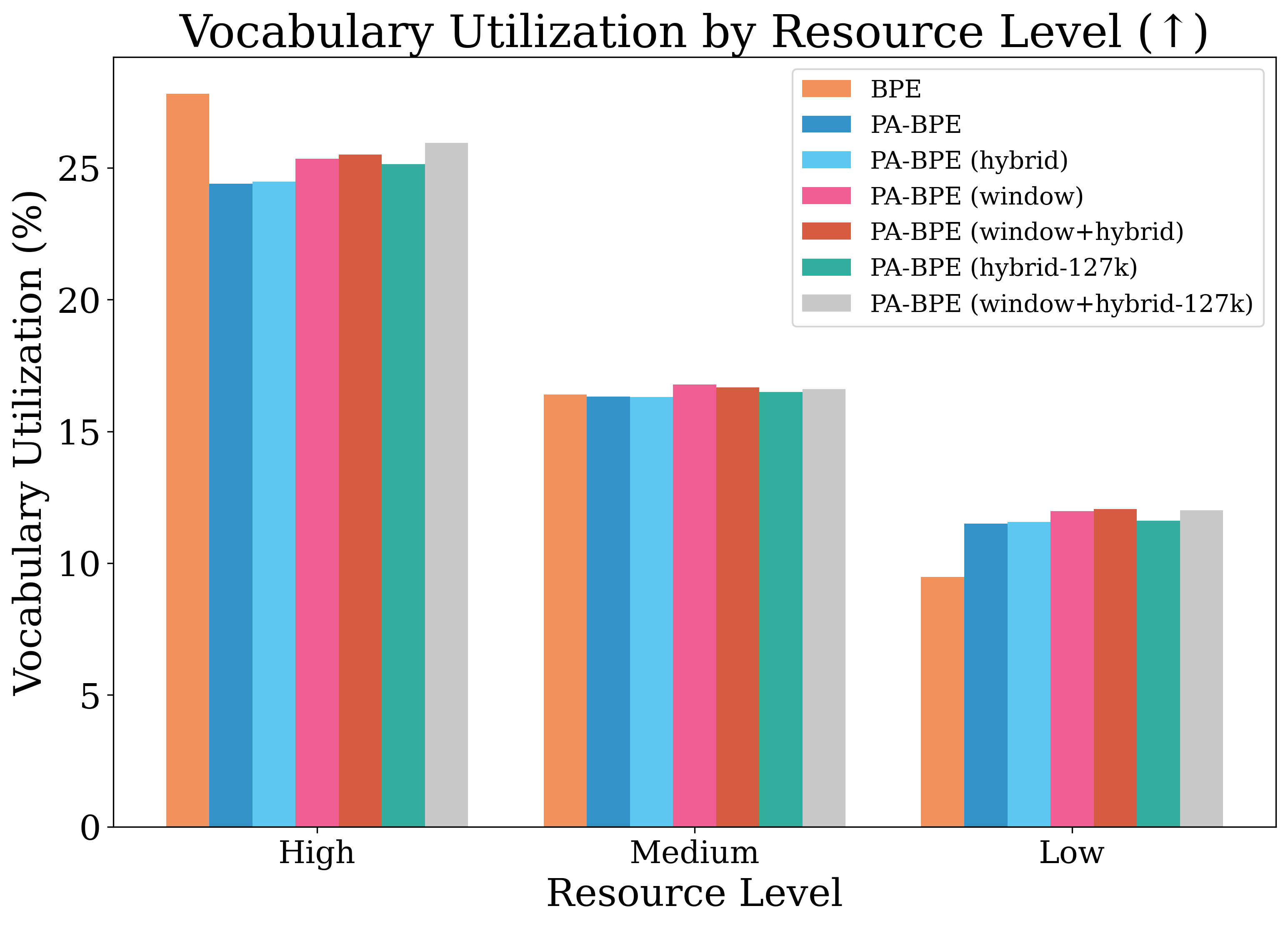}
        
    \end{subfigure}
    \hfill
    \begin{subfigure}[t]{0.49\textwidth}
        \centering
        \includegraphics[width=\linewidth]{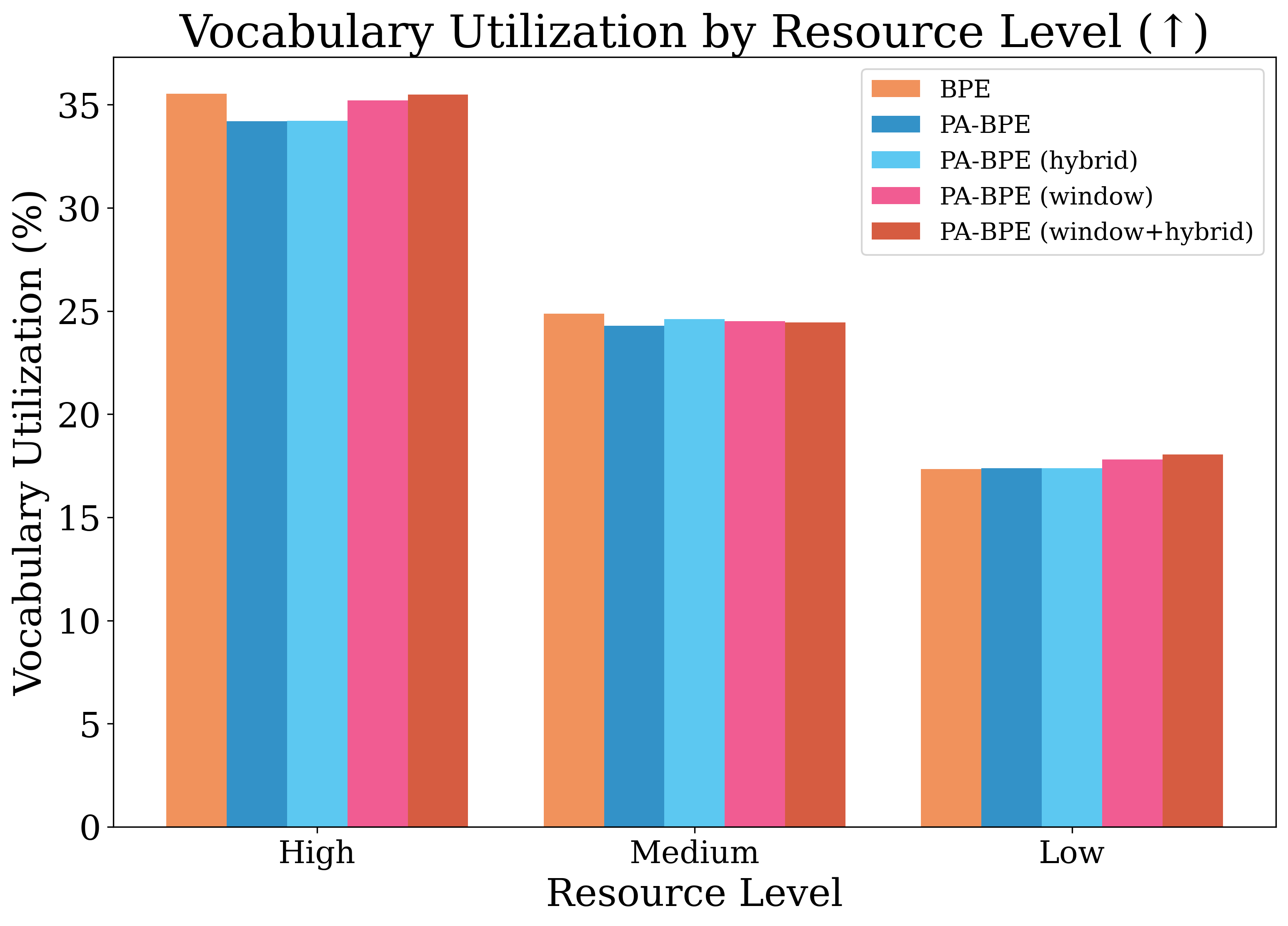}
        
    \end{subfigure}
    \caption{Vocabulary utilization grouped by language resource levels for the $256k$ tokenizer trained on the (\unbalanced) \thirtylangs{} mC4 dataset (left) and $128k$  tokenizer trained on the (\balanced) mC4 \thirtylangs{} dataset (right).}\label{fig:vocab_util_resource_level_balanced_30langs_128k}\label{fig:vocab_util_resource_level_unbalanced_30langs_256k}
\end{figure*}

\paragraph{Training Data Distribution.}
To assess the sensitivity of the Parity-aware algorithm to training data distribution, we also analyze results for $128k$ tokenizers trained on the \balanced version of the dataset. 
The results in \cref{tab:full_balanced_30langs_128k} indicate that Parity-aware BPE tokenizers perform similarly to Classical BPE across most metrics. However, in terms of fertility, Classical BPE outperforms the Parity-aware variants. This suggests that Parity-aware tokenizers are particularly beneficial in unbalanced settings, where low-resource languages are more disadvantaged, whereas in balanced scenarios their advantage diminishes.
We also interestingly see in \cref{fig:vocab_util_resource_level_balanced_30langs_128k}---again for the (\balanced) \thirtylangs{} setting---that Parity-aware tokenizers yield the largest absolute increases in vocabulary utilization for high-resource languages. Low- and medium-resource languages also improve, though to a smaller extent. One logical conclusion from this result is that the effect of Parity-aware tokenizer is better described as balancing utilization across languages rather than directly compensating for data scarcity.

\begin{table*}[htbp]
\adjustbox{max width=\textwidth}{%
\centering

\begin{tabular}{lcccccccc}
\toprule
Tokenizer & \makecell{Comp. Rate \\ ($\uparrow$)} & \makecell{TTR \\ ($\uparrow$)} & \makecell{Vocab \\ Utilization \\ ($\uparrow$)} & \makecell{Fertility \\ ($\downarrow$)} & \makecell{Rényi \\ ($\alpha$=2.5) \\ ($\uparrow$)} & \makecell{Gini \\ ($\downarrow$)} & \makecell{MorphScore \\ Precision \\ ($\uparrow$)} & \makecell{MorphScore \\ Recall \\ ($\uparrow$)} \\
\midrule
BPE & 0.0262 & 0.0718 & 65.1\% & 4.178 {\small $\pm$ 0.048} & \textbf{0.50} & 0.067 & \textbf{0.434 {\small $\pm$ 0.045}} & \textbf{0.527 {\small $\pm$ 0.042}} \\
UnigramLM & 0.0193 & 0.0457 & 56.3\% & 4.536 {\small $\pm$ 0.041} & 0.29 & 0.095 & 0.147 {\small $\pm$ 0.035} & 0.262 {\small $\pm$ 0.049} \\
PA-BPE & 0.0260 & 0.0727 & 66.4\% & 4.138 {\small $\pm$ 0.047} & 0.50 & \textbf{0.027} & 0.418 {\small $\pm$ 0.048} & 0.508 {\small $\pm$ 0.044} \\
PA-BPE (hybrid) & 0.0264 & 0.0737 & 66.3\% & \textbf{4.116 {\small $\pm$ 0.048}} & 0.50 & 0.027 & 0.426 {\small $\pm$ 0.047} & 0.515 {\small $\pm$ 0.044} \\
PA-BPE (window) & 0.0262 & 0.0749 & 67.9\% & 4.152 {\small $\pm$ 0.048} & 0.50 & 0.029 & 0.417 {\small $\pm$ 0.046} & 0.507 {\small $\pm$ 0.042} \\
PA-BPE (window+hybrid) & \textbf{0.0266} & \textbf{0.0761} & \textbf{67.9\%} & 4.127 {\small $\pm$ 0.048} & 0.49 & 0.031 & 0.430 {\small $\pm$ 0.045} & 0.517 {\small $\pm$ 0.043} \\
\bottomrule
\end{tabular}

}
\caption{Intrinsic evaluation of $128k$ tokenizers on the (\unbalanced) \textbf{\thirtylangs{}} mC4 dataset. Values are global statistics, except for MorphScore, which is macro-averaged across available languages. }
\label{tab:full_unbalanced_30langs_128k_mc4}
\end{table*}

\begin{table*}[htbp]
\adjustbox{max width=\textwidth}{%
\centering

\begin{tabular}{lcccccccc}
\toprule
Tokenizer & \makecell{Comp. Rate \\ ($\uparrow$)} & \makecell{TTR \\ ($\uparrow$)} & \makecell{Vocab \\ Utilization \\ ($\uparrow$)} & \makecell{Fertility \\ ($\downarrow$)} & \makecell{Rényi \\ ($\alpha$=2.5) \\ ($\uparrow$)} & \makecell{Gini \\ ($\downarrow$)} & \makecell{MorphScore \\ Precision \\ ($\uparrow$)} & \makecell{MorphScore \\ Recall \\ ($\uparrow$)} \\
\midrule
BPE & 0.0222 & \textbf{0.0341} & \textbf{72.8\%} & \textbf{3.564 {\small $\pm$ 0.026}} & 0.51 & 0.144 & \textbf{0.335 {\small $\pm$ 0.029}} & \textbf{0.481 {\small $\pm$ 0.027}} \\
PA-BPE & 0.0212 & 0.0264 & 58.9\% & 3.707 {\small $\pm$ 0.027} & \textbf{0.52} & \textbf{0.083} & 0.288 {\small $\pm$ 0.027} & 0.466 {\small $\pm$ 0.026} \\
PA-BPE (hybrid) & 0.0219 & 0.0279 & 60.4\% & 3.624 {\small $\pm$ 0.027} & 0.52 & 0.100 & 0.293 {\small $\pm$ 0.027} & 0.450 {\small $\pm$ 0.027} \\
PA-BPE (window) & 0.0218 & 0.0321 & 70.1\% & 3.647 {\small $\pm$ 0.027} & 0.51 & 0.101 & 0.314 {\small $\pm$ 0.027} & 0.469 {\small $\pm$ 0.026} \\
PA-BPE (window+hybrid) & \textbf{0.0222} & 0.0338 & 72.1\% & 3.576 {\small $\pm$ 0.027} & 0.50 & 0.116 & 0.323 {\small $\pm$ 0.027} & 0.466 {\small $\pm$ 0.027} \\
\bottomrule
\end{tabular}

}
\caption{Intrinsic evaluation of $128k$ tokenizers on the (\unbalanced) \textbf{\sixtylangs{}} mC4 dataset. Values are global statistics, except for MorphScore, which is macro-averaged across available languages. }
\label{tab:full_unbalanced_60langs_128k}
\end{table*}

\begin{table*}[htbp]
\adjustbox{max width=\textwidth}{%
\centering

\begin{tabular}{lcccccccc}
\toprule
Tokenizer & \makecell{Comp. Rate \\ ($\uparrow$)} & \makecell{TTR \\ ($\uparrow$)} & \makecell{Vocab \\ Utilization \\ ($\uparrow$)} & \makecell{Fertility \\ ($\downarrow$)} & \makecell{Rényi \\ ($\alpha$=2.5) \\ ($\uparrow$)} & \makecell{Gini \\ ($\downarrow$)} & \makecell{MorphScore \\ Precision \\ ($\uparrow$)} & \makecell{MorphScore \\ Recall \\ ($\uparrow$)} \\
\midrule
BPE & \textbf{0.0267} & 0.0758 & 67.4\% & \textbf{4.092} & 0.49 & 0.051 & \textbf{0.429} & 0.513 \\
PA-BPE & 0.0259 & 0.0726 & 66.4\% & 4.141 & \textbf{0.50} & 0.027 & 0.416 & 0.508 \\
PA-BPE (hybrid) & 0.0264 & 0.0738 & 66.3\% & 4.113 & 0.50 & \textbf{0.023} & 0.420 & 0.508 \\
PA-BPE (window) & 0.0261 & 0.0748 & 67.9\% & 4.155 & 0.50 & 0.029 & 0.418 & 0.507 \\
PA-BPE (window+hybrid) & 0.0266 & \textbf{0.0764} & \textbf{68.1\%} & 4.098 & 0.49 & 0.027 & 0.429 & \textbf{0.513} \\
\bottomrule
\end{tabular}

}
\caption{Intrinsic evaluation of $128k$ tokenizers on the (\textcolor{OliveGreen}{\textit{balanced}}) \textbf{\thirtylangs{}} mC4 dataset.
Values are global statistics, except for MorphScore, which is macro-averaged across available languages.
}
\label{tab:full_balanced_30langs_128k}
\end{table*}

\begin{table*}[htbp]
\adjustbox{max width=\textwidth}{%
\centering

\begin{tabular}{lcccccccc}
\toprule
Tokenizer & \makecell{Comp. Rate \\ ($\uparrow$)} & \makecell{TTR \\ ($\uparrow$)} & \makecell{Vocab \\ Utilization \\ ($\uparrow$)} & \makecell{Fertility \\ ($\downarrow$)} & \makecell{Rényi \\ ($\alpha$=2.5) \\ ($\uparrow$)} & \makecell{Gini \\ ($\downarrow$)} & \makecell{MorphScore \\ Precision \\ ($\uparrow$)} & \makecell{MorphScore \\ Recall \\ ($\uparrow$)} \\
\midrule
BPE & 0.0294 & 0.1200 & 48.3\% & \textbf{3.696 {\small $\pm$ 0.042}} & 0.46 & 0.056 & 0.529 {\small $\pm$ 0.047} & \textbf{0.598 {\small $\pm$ 0.044}} \\
PA-BPE & 0.0288 & 0.1143 & 47.0\% & 3.752 {\small $\pm$ 0.043} & \textbf{0.47} & 0.028 & 0.501 {\small $\pm$ 0.050} & 0.570 {\small $\pm$ 0.046} \\
PA-BPE (hybrid) & 0.0291 & 0.1157 & 47.1\% & 3.739 {\small $\pm$ 0.043} & 0.47 & \textbf{0.026} & 0.513 {\small $\pm$ 0.049} & 0.582 {\small $\pm$ 0.046} \\
PA-BPE (window) & 0.0292 & 0.1203 & 48.9\% & 3.725 {\small $\pm$ 0.043} & 0.47 & 0.032 & 0.514 {\small $\pm$ 0.049} & 0.580 {\small $\pm$ 0.045} \\
PA-BPE (window+hybrid) & 0.0294 & 0.1217 & 49.0\% & 3.709 {\small $\pm$ 0.043} & 0.46 & 0.030 & 0.524 {\small $\pm$ 0.048} & 0.590 {\small $\pm$ 0.045} \\
PA-BPE (hybrid-127k) & 0.0294 & 0.1188 & 47.9\% & 3.722 {\small $\pm$ 0.043} & 0.46 & 0.027 & 0.521 {\small $\pm$ 0.049} & 0.590 {\small $\pm$ 0.045} \\
PA-BPE (window+hybrid-127k) & \textbf{0.0296} & \textbf{0.1231} & \textbf{49.3\%} & 3.698 {\small $\pm$ 0.043} & 0.46 & 0.032 & \textbf{0.530 {\small $\pm$ 0.048}} & 0.596 {\small $\pm$ 0.045} \\
\bottomrule
\end{tabular}

}
\caption{Intrinsic evaluation of $256k$ tokenizers on the (\textcolor{BrickRed}{\textit{unbalanced}}) \textbf{\thirtylangs{}} mC4 dataset.
Values are global statistics, except for MorphScore, which is macro-averaged across available languages.
}
\label{tab:full_unbalanced_30langs_256k}
\end{table*}

\paragraph{Language Model Perplexities.}
We report language model perplexities on the \finewebtwo{} validation set in \cref{fig:per-lang-perp}. Results are shown per language. We see a noticeably larger cross-lingual spread in perplexity for language models trained using the Classical BPE tokenizer than for those trained using Parity-aware variants. The Parity-aware tokenizers seem to eliminate the long tail present under Classical BPE while maintaining comparable mean perplexity across languages. Note that we normalize by number of bytes in the text rather than by number of tokens to account for differences in tokenization lengths.
\begin{figure*}
    \includesvg[width=\linewidth]{figs/128k/perplexity.svg}
    \caption{Per-language perplexities ($\downarrow$) normalized by byte of language models trained using the specified tokenizer trained on the (\unbalanced) $128k$ \thirtylangs{} dataset. Results are computed on the language model validation set at the final checkpoint (see \cref{appendix:training_setup} for language model details). }\label{fig:per-lang-perp}
\end{figure*}

\section{Balancing the tokenizer training set}
\label{app:balancing_training_data}

Prior work has addressed representational imbalance in subword tokenization primarily through corpus-level reweighting or balancing strategies, such as equalizing the number of documents or bytes per language prior to training a BPE or unigram tokenizer. While such approaches ensure that low-resource languages contribute to the learned vocabulary, they optimize the tokenizer with respect to an artificial distribution that differs from the natural data distribution encountered during downstream training and inference. This mismatch can result in inefficient allocation of vocabulary capacity, over-representation of rare patterns, and increased tokenization length for high-frequency languages. Moreover, corpus balancing constitutes a coarse-grained intervention: the choice of balancing criterion substantially affects outcomes, and the approach does not account for heterogeneity within groups or for the local utility of individual merge operations.

Parity-aware BPE instead operates on the unaltered corpus distribution and introduces language-level constraints directly into the merge selection process. By augmenting merge scores with Parity-aware penalties or budgets, this approach enables fine-grained control over how subword units are allocated across languages, while preserving the efficiency benefits of frequency-driven merging. In contrast to the balancing approach, Parity-aware merges maintain alignment with the downstream data distribution and typically yield lower average tokenization costs at inference, while mitigating systematic over-fragmentation of underrepresented languages. Overall, Parity-aware BPE offers a principled mechanism for addressing tokenizer bias at the level of individual merge decisions, complementing existing corpus-level mitigation strategies.
\section{Language Model Training}
\label{appendix:training_setup}
Here we provide details about the language models used for evaluating extrinsic tokenizer metrics. 
\subsection{Model Architecture}
Our experiments focus on models with 3 billion parameters ($3$B), following the LLaMA architecture~\citep{touvron2023llama}.
Model size is determined by adjusting the number of layers, hidden dimensions, and attention heads.

\subsection{Training Hyperparameters}
We train our models using SwissAI's fork of Hugging Face's Nanotron repository.\footnote{The codebase: \href{https://github.com/swiss-ai/nanotron-multilingual}{https://github.com/swiss-ai/nanotron-multilingual}}
The key training hyperparameters are as follows:
\begin{itemize}[leftmargin=*,noitemsep]
    \item \textbf{Learning Rate.} We use a learning rate of $8 \times 10^{-4}$ with linear warmup over the first 4\% of training. A ``1-sqrt'' decay schedule~\citep{hagele2024scaling} is applied during the final 20\%.
    \item \textbf{Optimizer.} All experiments use AdamW with $\beta = (0.9, 0.95)$~\citep{loshchilov2017decoupled}.
    \item \textbf{Weight Decay.} We set the weight decay parameter to $\lambda = 0.1$ for regularization.
    \item \textbf{Batch Size.} The micro-batch size is fixed at 5 across all runs.
\end{itemize}

\subsection{Hardware Setup}
Training is performed on a large-scale cluster. Each node is equipped with 4 NVIDIA Grace-Hopper H100 GPUs (96 GB memory each).
The 3B models are trained on 64 nodes (256 GPUs in total), requiring approximately 18 hours per 100B tokens.
This corresponds to a global batch size of 640 examples.

\subsection{Sampling Methods}
\label{appendix:sampling_methods}
Let $\mathcal{L}$ be the set of languages in the dataset, and let $\pi^{\text{natural}} \in \Delta_{|\mathcal{L}|}$ represent the natural distribution of these languages, defined as:
\[
\pi_l^{\text{natural}} = \frac{\omega_l}{\sum_{l' \in \mathcal{L}} \omega_{l'}}
\]
where $\omega_l$ denotes the number of words (or tokens) for language $l$ in the dataset. In this work, we use the number of words as a proxy for language frequency, a common practice when presenting statistics for highly multilingual datasets~\cite{penedo2025fineweb}.
In order to create our LM training dataset, we use temperature sampling. This method adjusts the natural distribution using a temperature parameter $\tau$ to create a less skewed distribution:
    \[
    \pi_l^{\text{temp}, \tau} = \frac{\omega_l^{1/\tau}}{\sum_{l' \in \mathcal{L}} \omega_{l'}^{1/\tau}}
    \]
    By tuning $\tau$, the distribution can be shifted towards uniformity, thereby reducing imbalance among languages.

\begin{table*}[h]
\centering
\resizebox{0.85\textwidth}{!}{
\begin{tabular}{lll c c c}
\toprule
\textbf{Language} & \textbf{Language Family} & \textbf{Script} &\textbf{Resource Level}  & \textbf{30-lang} & \textbf{60-lang}\\
\midrule
English          & Indo-European (Germanic) & Latin & High & \checkmark & \checkmark\\
German           & Indo-European (Germanic) & Latin & High & \checkmark & \checkmark\\
French           & Indo-European (Romance) & Latin & High & \checkmark & \checkmark\\
Italian          & Indo-European (Romance) & Latin & High & \checkmark & \checkmark\\
Russian          & Indo-European (Slavic) & Cyrillic & High & \checkmark & \checkmark\\
Spanish          & Indo-European (Romance) & Latin & High & \checkmark & \checkmark\\
Japanese         & Japonic & Kanji \& Kana (CJK) & Medium & \checkmark & \checkmark\\
Polish           & Indo-European (Slavic) & Latin &Medium &  \checkmark & \checkmark\\
Portuguese       & Indo-European (Romance) & Latin &Medium &\checkmark & \checkmark \\
Vietnamese       & Austroasiatic &  Latin &Medium & \checkmark & \checkmark \\
Turkish          & Turkic &  Latin & Medium & \checkmark & \checkmark \\
Dutch            & Indo-European (Germanic) & Latin & High & \checkmark & \checkmark \\
Indonesian       & Austronesian  & Latin  & Medium & \checkmark & \checkmark \\
Arabic           & Afro-Asiatic (Semitic) & Perso-Arabic & Medium & \checkmark & \checkmark \\
Czech            & Indo-European (Slavic) & Latin &Medium & \checkmark & \checkmark \\
Persian (Farsi)  & Indo-European (Iranian) & Perso-Arabic & Medium & \checkmark & \checkmark \\
Greek            & Indo-European (Hellenic) & Greek &Medium & \checkmark & \checkmark \\
Chinese (Mandarin) & Sino-Tibetan & Hanzi (CJK) &Medium & \checkmark & \checkmark \\
Hindi            & Indo-European (Indo-Aryan) &  Devanagari (Brahmic) &Medium  & \checkmark & \checkmark \\
Korean           & Koreanic & Hangugeo (CJK) & Medium & \checkmark & \checkmark \\
Thai             & Kra–Dai (Tai) & Thai & Medium & \checkmark & \checkmark \\
Hebrew           & Afro-Asiatic (Semitic) & Hebrew & Medium & \checkmark & \checkmark \\
Bengali          & Indo-European (Indo-Aryan) & Bengali (Brahmic) &Medium & \checkmark & \checkmark \\
Tamil            & Dravidian (Brahmic) & Tamil & Low & \checkmark & \checkmark \\
Georgian         & Kartvelian & Georgian & Low & \checkmark & \checkmark \\
Marathi          & Indo-European (Indo-Aryan) & Devanagari (Brahmic) & Medium & \checkmark & \checkmark \\
Filipino         & Austronesian & Latin & Low & \checkmark & \checkmark \\
Telugu           & Dravidian  & Telugu (Brahmic) & Low & \checkmark & \checkmark \\
Norwegian        & Indo-European (Germanic) & Latin & Medium & \checkmark & \checkmark \\
North Azerbaijani& Turkic & Latin &  Low & \checkmark & \checkmark \\
Swedish         & Indo-European (Germanic)  & Latin & Medium & - & \checkmark \\
Romanian        & Indo-European (Romance) & Latin & Medium & - & \checkmark \\
Ukrainian       & Indo-European (Slavic) & Cyrillic & Medium & - & \checkmark \\
Hungarian       & Uralic (Ugric) & Latin & Medium & - & \checkmark \\
Danish          & Indo-European (Germanic) & Latin & Medium & - & \checkmark \\
Finnish         & Uralic (Finnic) & Latin & Medium & - & \checkmark \\
Bulgarian       & Indo-European (Slavic) & Cyrillic & Low & - & \checkmark \\
Slovak          & Indo-European (Slavic) & Latin & Low & - & \checkmark \\
Catalan         & Indo-European (Romance) & Latin & Low & - & \checkmark \\
Malay           & Austronesian & Latin & Low & - & \checkmark \\
Urdu            & Indo-European (Indo-Aryan) & Perso-Arabic & Low & - & \checkmark \\
Belarusian      & Indo-European (Slavic) & Cyrillic & Medium & - & \checkmark \\
Basque          & Language Isolate  & Latin & Low & - & \checkmark \\
Tajik           & Indo-European (Iranian) & Cyrillic & Medium & - & \checkmark \\
Sotho (Sesotho) & Niger–Congo (Bantu)& Latin & Low & - & \checkmark \\
Yoruba          & Niger–Congo & Latin & Low & - & \checkmark \\
Swahili         & Niger-Congo (Bantu) & Latin & Low & - & \checkmark \\
Estonian        & Uralic (Finnic)  & Latin & Low & - & \checkmark \\
Latvian         & Indo-European (Slavic) & Latin & Low & - & \checkmark \\
Galician        & Indo-European (Romance) & Latin & Low & - & \checkmark \\
Welsh           & Indo-European (Celtic)  & Latin & Low & - & \checkmark \\
Albanian        & Indo-European & Latin & Low & - & \checkmark \\
Macedonian      & Indo-European (Slavic) & Cyrillic & Low & - & \checkmark \\
Malayalam       & Dravidian & Malayalam (Brahmic) & Low & - & \checkmark \\
Burmese         & Sino-Tibetan & Mon–Burmese & Low & - & \checkmark \\
Gujarati        & Indo-European (Indo-Aryan) & Gujarati (Brahmic)& Low & - & \checkmark \\
Afrikaans       & Indo-European (Germanic) & Latin & Low & - & \checkmark \\
Hawaiian        & Austronesian & Latin & Low & - & \checkmark \\
Northern Uzbek  & Turkic & Latin & Low & - & \checkmark \\

\bottomrule
\end{tabular}
}
\caption{Details on the languages used to train and evaluate tokenizers.}
\label{tab:language_list}
\end{table*}

\section{Downstream Benchmark Evaluation}

\label{appendix:benchmark_setup}
We evaluate our models using HuggingFace's Lighteval codebase~\cite{lighteval}.

\subsection{Benchmarks}
To assess multilingual performance, we select twelve widely used benchmarks that span diverse downstream tasks, including reading comprehension, commonsense reasoning, semantic similarity, and knowledge-based evaluation~\citep{ustun2024aya, swissai2025apertus, martins2025eurollm}. At the same time, we ensured that the chosen benchmarks provide meaningful performance signals (results clearly above random chance) for small-scale models (1-3B parameters). Additionally, all selected benchmarks can be evaluated in a zero-shot setting and do not require supervised finetuning.
To ensure consistency with prior work, we followed established evaluation practices.

\begin{itemize}
\item \textbf{Belebele}:\footnote{\url{https://huggingface.co/datasets/facebook/belebele}} A multilingual reading comprehension dataset containing passages and corresponding questions in many languages. It evaluates models’ ability to understand text and answer related questions~\citep{bandarkar-etal-2024-belebele}.
    \item \textbf{mTruthfulQA}:\footnote{\url{https://huggingface.co/datasets/alexandrainst/m_truthfulqa}} A multilingual extension of TruthfulQA, designed to measure whether models generate accurate and non-misleading answers across a broad range of questions~\citep{lin-etal-2022-truthfulqa, dac2023okapi}.
    \item \textbf{PAWS-X}:\footnote{\url{https://huggingface.co/datasets/google-research-datasets/paws-x}} A multilingual paraphrase identification benchmark that extends the original PAWS dataset, providing sentence pairs annotated for semantic equivalence~\citep{yang-etal-2019-paws}.
    \item \textbf{XCodah}:\footnote{\url{https://huggingface.co/datasets/INK-USC/xcsr}} A multilingual adaptation of CODAH for adversarially-authored commonsense reasoning tasks, testing robustness in natural language understanding~\citep{lin-etal-2021-common, Chen2019CODAHAA}.
    \item \textbf{XCSQA}:\footnote{\url{https://huggingface.co/datasets/INK-USC/xcsr}}  A multilingual version of CommonsenseQA, consisting of multiple-choice questions that require reasoning about everyday concepts and their relations~\citep{lin-etal-2021-common, Talmor2019commonsenseqaaq}.
    \item \textbf{XNLI}:\footnote{\url{https://huggingface.co/datasets/facebook/xnli}} A cross-lingual natural language inference benchmark, evaluating whether models can perform entailment, contradiction, and neutrality classification across multiple languages~\citep{conneau-etal-2018-xnli}.
    \item \textbf{XStoryCloze}:\footnote{\url{https://huggingface.co/datasets/juletxara/xstory_cloze}} A multilingual extension of the StoryCloze Test, where models must choose the most coherent ending to short narratives, testing story comprehension and commonsense reasoning~\citep{mostafazadeh2017lsdsem, xi2022xstorycloze}.
    \item \textbf{XWinogrande}:\footnote{\url{https://huggingface.co/datasets/allenai/winogrande}} A multilingual version of WinoGrande, containing sentences with ambiguous pronouns. It measures models’ ability to resolve coreference using contextual and commonsense cues~\citep{ai2:winogrande, muennighoff2022crosslingual, tikhonov2021heads}.
    \item \textbf{MMMLU}:\footnote{\url{https://huggingface.co/datasets/openai/MMMLU}} A multilingual adaptation of MMLU, evaluating model performance across a wide spectrum of tasks and domains~\citep{hendryckstest2021, dac2023okapi}.
    \item \textbf{INCLUDE}:\footnote{\url{https://huggingface.co/datasets/CohereLabs/include-base-44}} A large-scale benchmark covering 44 languages, designed to evaluate multilingual LLMs in realistic language environments with a focus on knowledge and reasoning~\citep{romanou2024include}.
    \item \textbf{Exams}:\footnote{\url{https://huggingface.co/datasets/mhardalov/exams}} A benchmark of standardized test questions across subjects and educational levels, used to assess reasoning and problem-solving abilities in exam-like conditions~\citep{hardalov-etal-2020-exams}.
    \item \textbf{M3Exams}:\footnote{\url{https://huggingface.co/datasets/SEACrowd/m3exam}} A multilingual exam-style benchmark that extends Exams across different languages, subjects, and difficulty levels~\citep{zhang2023m3exam}.
\end{itemize}

\subsection{Score Aggregations}
We aggregate benchmark results to compute a language-specific score for each model. Let $\mathcal{T}_l$ be the set of benchmarks (or tasks) containing a split for language $l$. The aggregated score for a model $m$ per language $l$ is defined as:
\[
s_l^m = \frac{1}{|\mathcal{T}_l|}\sum_{t \in \mathcal{T}_l} s_{t,l}^m
\]
where $s_l^m$ is the score of a model $m$ on the split $l$ of a task $t$
To mitigate biases arising from varying numbers of benchmarks per language, we compute a language-specific random baseline $\zeta_l$. This baseline helps assess whether a given aggregated score significantly outperforms random predictions. Specifically, we calculate the random baseline for each language as the average of the individual random baselines across all tasks that include language $l$:
\[
\zeta_l = \frac{1}{|\mathcal{T}_l|}\sum_{t \in \mathcal{T}_l} \zeta_t
\]

\begin{table*}
\centering
\resizebox{\textwidth}{!}{%
\begin{tabular}{lcccccccccccc}
\toprule
\textbf{Language} & INCLUDE & Belebele & Exams & M3Exam & MMMLU & mTruthfulQA & PAWS-X & XCodah & XCSQA & XNLI & XStoryCloze & XWinoGrande \\
\midrule
English &  - & \checkmark &  - & \checkmark & \checkmark & \checkmark & \checkmark & \checkmark & \checkmark & \checkmark & \checkmark & \checkmark \\
Chinese & \checkmark & \checkmark &  - & \checkmark & \checkmark & \checkmark & \checkmark & \checkmark & \checkmark & \checkmark & \checkmark & \checkmark \\
Vietnamese & \checkmark & \checkmark & \checkmark & \checkmark & \checkmark & \checkmark &  - & \checkmark & \checkmark & \checkmark &  - & - \\
Arabic & \checkmark & \checkmark & \checkmark &  - & \checkmark & \checkmark &  - & \checkmark & \checkmark & \checkmark & \checkmark & - \\
German & \checkmark & \checkmark & \checkmark &  - & \checkmark & \checkmark & \checkmark & \checkmark & \checkmark &  - & -  & - \\
Spanish & \checkmark & \checkmark & \checkmark &  - & \checkmark & \checkmark &  - & \checkmark & \checkmark & \checkmark & \checkmark & - \\
French & \checkmark & \checkmark &  - &  - & \checkmark & \checkmark & \checkmark & \checkmark & \checkmark & \checkmark &  - & \checkmark \\
Portuguese & \checkmark & \checkmark & \checkmark &  - & \checkmark & \checkmark &  - & \checkmark & \checkmark &  - &  - & \checkmark \\
Hindi & \checkmark & \checkmark &  - &  - & \checkmark & \checkmark &  - & \checkmark & \checkmark & \checkmark & \checkmark & - \\
Russian & \checkmark & \checkmark &  - &  - & \checkmark & \checkmark &  - & \checkmark & \checkmark & \checkmark & \checkmark & \checkmark \\
Indonesian & \checkmark &  - &  - &  - & \checkmark & \checkmark &  - &  - &  - & -  & \checkmark & - \\
Italian & \checkmark & \checkmark & \checkmark & \checkmark & \checkmark & \checkmark &  - & \checkmark & \checkmark &  - &  - & - \\
Japanese & \checkmark &  - &  - &  - &  - &  - & \checkmark & \checkmark & \checkmark &  - & -  & \checkmark \\
Swahili &  - & \checkmark &  - & \checkmark &  - &  - &  - & \checkmark & \checkmark & \checkmark & \checkmark & - \\
Tamil & \checkmark &  - &  - &  - &  - &  - &  - &  - &  - &  - &  - & - \\
Telugu & \checkmark & \checkmark &  - &  - & \checkmark & \checkmark &  - &  - &  - &  - & \checkmark & - \\
Thai &  - & \checkmark &  - & \checkmark &  - &  - &  - &  - &  - & \checkmark &  - & - \\
Basque & \checkmark &  - &  - &  - & \checkmark & \checkmark &  - &  - &  - &  - & \checkmark & - \\
Turkish & \checkmark & \checkmark & \checkmark &  - &  - &  - &  - &  - &  - & \checkmark &  - & - \\
Bulgarian & \checkmark & \checkmark & \checkmark &  - &  - &  - &  - &  - &  - & \checkmark &  - & - \\
Albanian & \checkmark & \checkmark & \checkmark &  - &  - &  - &  - &  - &  - &  - &  - & - \\
Polish & \checkmark & \checkmark &  - &  - &  - &  - &  - & \checkmark &  - &  - &  - & - \\
Bengali & \checkmark &  - &  - &  - & \checkmark & \checkmark &  - &  - &  - &  - &  - & - \\
Serbian & \checkmark &  - & \checkmark &  - &  - & \checkmark &  - &  - &  - &  - &  - & - \\
Estonian & \checkmark &  - &  - &  - &  - &  - &  - &  - &  - &  - &  - & - \\
Macedonian & \checkmark & \checkmark &  - &  - &  - &  - &  - &  - &  - &  - &  - & - \\
Lithuanian & \checkmark &  - & \checkmark &  - &  - &  - &  - &  - &  - &  - &  - & - \\
Greek & \checkmark &  - &  - &  - &  - &  - &  - &  - &  - & \checkmark &  - & - \\
Urdu & \checkmark &  - &  - &  - &  - &  - &  - &  - &  - & \checkmark &  - & - \\
Catalan &  - &  - &  - &  - & \checkmark & \checkmark &  - & -  &  - &  - &  - & - \\
Persian & \checkmark &  - &  - &  - &  - &  - &  - &  - &  - &  - &  - & - \\
Finnish & \checkmark &  - &  - &  - &  - &  - &  - &  - &  - &  - &  - & - \\
Korean & \checkmark &  - &  - &  - &  - &  - &  - &  - &  - &  - &  - & - \\
Quechua &  - &  - &  - &  - &  - &  - &  - &  - &  - &  - &  - & - \\
Haitian Creole &  - &  - &  - &  - &  - &  - &  - &  - &  - &  - &  - & - \\
Malay &  - &  - &  - &  - &  - &  - &  - &  - &  - &  - & \checkmark & - \\
\bottomrule
\end{tabular}%
}
\caption{Coverage of downstream benchmarks across languages.}
\label{tab:language_benchmark-coverage}
\end{table*}

\end{document}